\documentclass[conference]{IEEEtran}
\IEEEoverridecommandlockouts
\usepackage{cite}
\usepackage{amsmath,amssymb,amsfonts}
\usepackage{graphicx}
\usepackage{textcomp}
\usepackage{xcolor}
\usepackage{multirow}
\usepackage{makecell}
\usepackage{subcaption}
\usepackage{minted}
\usepackage{utfsym}
\usepackage{listings}
\lstdefinelanguage{json}{
    sensitive=true,
    morekeywords={true,false,null},
    morestring=[b]"
}
\usepackage{bbm}
\usepackage{booktabs}
\usepackage{array}
\usepackage{booktabs}
\usepackage{multirow}
\usepackage{pifont}
\usepackage{enumitem}
\usepackage{tabularx}
\usepackage{hyperref}
\setlist[itemize]{leftmargin=2em}
\setlist[enumerate]{leftmargin=*}
\def\BibTeX{{\rm B\kern-.05em{\sc i\kern-.025em b}\kern-.08em
    T\kern-.1667em\lower.7ex\hbox{E}\kern-.125emX}}
\newcommand{\cmark}{\textcolor{olive}{\ding{51}}} 
\newcommand{\xmark}{\textcolor{red}{\ding{55}}}
\usepackage{amsmath}
\usepackage{amssymb}
\usepackage{mathtools}
\usepackage{amsthm}
\usepackage{booktabs}
\usepackage[table]{xcolor}
\usepackage{multirow}
\usepackage[most]{tcolorbox}
\usepackage{listings}
\usepackage{xcolor}
\usepackage{enumitem}
\usepackage{tikz}
\usepackage{pgfplots}
\pgfplotsset{compat=1.18} 
\usepackage{xcolor}      
\usepackage[T1]{fontenc} 
\usepackage{orcidlink}
\usepackage{diagbox}

\usepackage{algorithm}
\usepackage{algpseudocode}
\usepackage{caption}

\tcbuselibrary{skins, breakable}
\newtcolorbox{promptbox}[2][]{
  colback=gray!5!white,
  colframe=gray!75!black,
  title={#2},
  fonttitle=\bfseries,
  breakable,
  enhanced,
  boxrule=0.8pt,
  sharp corners,
  #1
}

\usepackage[margin=1in]{geometry}

\lstdefinestyle{pythonstyle}{
    language=Python,
    basicstyle=\ttfamily\small,
    keywordstyle=\color{blue}\bfseries,
    commentstyle=\color{green!60!black},
    stringstyle=\color{red},
    numbers=left,
    numberstyle=\tiny\color{gray},
    stepnumber=1,
    numbersep=5pt,
    backgroundcolor=\color{gray!10},
    showspaces=false,
    showstringspaces=false,
    showtabs=false,
    frame=single,
    rulecolor=\color{black!30},
    tabsize=4,
    captionpos=b,
    breaklines=true,
    breakatwhitespace=true,
    escapeinside={\%*}{*)},
    morekeywords={import, as, def, return, class, self, if, elif, else, for, while, try, except, lambda, with, yield, pd, np, os}
}
\newcommand{\method}{LYNX}
\newcommand{\temperature}{0}

\begin{document}

\title{LYNX: Analytical Program Synthesis via Grounding and Closed-Loop Refinement
}

\author{
\IEEEauthorblockN{
Ruichen Xia\textsuperscript{1,2}\IEEEauthorrefmark{1}\orcidlink{0009-0002-2736-451X},
Xiao Han\textsuperscript{1}\IEEEauthorrefmark{1}\orcidlink{0009-0003-5235-9592},
Dingyu Yang\textsuperscript{1}\IEEEauthorrefmark{2}\orcidlink{0000-0002-8156-3926},
Luwei Yang\textsuperscript{3}\orcidlink{0000-0003-3949-4119},\\
Sai Wu\textsuperscript{1}\orcidlink{0000-0002-7903-1496},
Dongxiang Zhang\textsuperscript{1}\orcidlink{0009-0006-6338-0698},
Binhui Yao\textsuperscript{1,3}\orcidlink{0000-0002-1998-8640},
Gang Chen\textsuperscript{1}\orcidlink{0000-0002-7483-0045}
}

\IEEEauthorblockA{
\textsuperscript{1}
The State Key Laboratory of Blockchain and Data Security,
Zhejiang University, Hangzhou, China\\
\textsuperscript{2}
Institute of Computing Innovation,
Zhejiang University, Hangzhou, China\\
\textsuperscript{3}
Zhejiang Geely Holding Group Co., Ltd., Hangzhou, China\\
rcxia@zju.edu.cn,
3230100653@zju.edu.cn,
yangdingyu@zju.edu.cn,
Luwei.Yang1@geely.com\\
wusai@zju.edu.cn,
zhangdongxiang@zju.edu.cn,
12321292@zju.edu.cn,
cg@zju.edu.cn
}

\thanks{\IEEEauthorrefmark{1} These authors contributed equally to this work.}
\thanks{\IEEEauthorrefmark{2} Corresponding author.}
}

\maketitle

\begin{abstract}

LLM-powered data analytics aims to translate natural language instructions into executable analytical programs over underlying data. However, existing approaches—whether treating LLMs as intent-to-code translators or as standalone analytical engines—often fail to reliably align domain semantics with complex schemas and heterogeneous metadata, resulting in brittle execution and semantically inconsistent analytical outputs in real-world workloads.
In this paper, we present {\method}, a multi-agent framework that formulates LLM-driven analytics as a \textbf{schema- and knowledge-constrained analysis} problem rather than a single-pass generation task. {\method} organizes analytical synthesis into a closed-loop workflow of planning, generation, verification, and refinement through three role-specialized agents that collaboratively enforce schema grounding, semantic consistency, and execution correctness. To address domain ambiguity and schema misalignment, {\method} integrates \textbf{Semantic-Structural Distillation and
Stage-Aware Allocation (SSD-SAA)}. Furthermore, a delta-guided iterative revision continuously diagnoses and repairs semantic and execution-level failures, substantially improving synthesis robustness.
We evaluate {\method} on two NL-to-SQL benchmarks, \textbf{BIRD-SQL Mini-Dev} and more challenging \textbf{Spider 2.0-Lite}. Experimental results demonstrate strong performance and efficiency, achieving 10.66\% to 59.54\% execution accuracy across lightweight and commercial models, while reducing token consumption by \textbf{70.53\%} compared with the prior most efficient method.
Also, our framework displays high generalization over different backbones, from open-source, lightweight models to the latest commercial models. Additional case studies on quantitative finance analytics and complex multi-relational SQL workloads further demonstrate the framework’s robustness, scalability, and practical applicability in real-world analytical scenarios. Our code is available at \url{https://github.com/summerhare12/LYNX}.

\end{abstract}


\section{Introduction}
\label{section::introduction}

LLM-powered data analytics aims to translate natural language analytical queries into executable analytical programs by jointly reasoning over query semantics, database schemas, metadata, and execution constraints~\cite{jiang2025siriusbi, minaee2024large, liu2024understanding}. This capability enables domain experts to express analytical intent through domain knowledge rather than low-level programming, making it particularly attractive for enterprise and scientific analytics involving complex workflows and heterogeneous datasets. However, reliably synthesizing executable analytical programs remains challenging in real-world environments with heterogeneous schemas, complex metadata, and domain-specific semantics. Even minor misalignment between natural language intent and structured data semantics can lead to execution failures or executable-but-incorrect analytical programs~\cite{wei2022emergent, guo2023evaluating, zhou2023dbgpt}.

Existing approaches to LLM-powered data analytics mainly follow two directions. The first treats LLMs as \emph{intent-to-code translators}, where models generate executable programs (e.g., SQL or Python) that are executed by external analytical engines. Representative systems include Text-to-SQL frameworks such as DIN-SQL~\cite{pourreza2023din}, CHESS~\cite{talaei2024chess}, and agent-based systems such as OpenAgents~\cite{xie2023openagents}. While these methods effectively leverage the code generation capability of LLMs, they typically rely on single-pass synthesis or static schema linking, making them vulnerable to weak schema grounding, execution failures, and semantically incorrect analytical programs when domain constraints are incompletely captured~\cite{wang2025towards, wang2024large}. 
The second direction employs LLMs as reasoning-centric analytical models that directly operate over tabular or structured data without explicit execution engines. Representative examples include Galois~\cite{saeed2023querying}, TableGPT~\cite{zha2023tablegpt}, and TableLLaMA~\cite{zhang2024tablellama}. Although these approaches reduce dependence on external systems, they often struggle with large-scale or multi-relational analytical workloads and are more susceptible to semantic, factual, and arithmetic reasoning errors compared with traditional database execution engines~\cite{zhang2025tablellm}.

Despite their architectural differences, both paradigms ultimately struggle with two fundamental issues: accurately grounding domain semantics to database schemas and ensuring semantic correctness throughout analytical program execution. To address these limitations, we identify the following fundamental challenges:

\textbf{C1: Domain and Schema Grounding.}
General-purpose LLMs often lack sufficient understanding of domain-specific terminology and the underlying database structures, including table schemas, attribute types, and inter-table relationships~\cite{cao2024spider2, zhang2023sciencebenchmark, liu2024survey, wang2025llm}. Consequently, models frequently produce incorrect schema linking, hallucinated attributes, inconsistent joins, or invalid type usage~\cite{wang2020rat, kothyari2023crush4sql, ma2010bridging, vaidya2025odin, pourreza2024sql}, leading to semantically incorrect or non-executable analytical programs.

\textbf{C2: Execution Reliability and Semantic Consistency.}
Even when generated programs are syntactically valid, LLMs often struggle to preserve semantic consistency across multi-step analytical reasoning, including proper function usage, condition composition, and dependency preservation~\cite{shen2025study, li2025hallucination, wang2025towards, wang2024large, xiong2024metamath}. 
Such limitations produce executable-but-incorrect programs that satisfy syntactic and execution constraints while violating the intended analytical semantics.
This highlights the inherent brittleness of single-pass analytical program synthesis and the need for explicit semantic verification.


To address these challenges, we propose {\method}, a framework that jointly integrates schema grounding, analytical reasoning, and semantic verification within a unified process. The framework consists of three coordinated components—\textbf{Generator}, \textbf{Critic}, and \textbf{Reviser}—that collaboratively perform grounded generation, execution-guided refinement, and intent-level validation.

To address \textbf{C1 (Domain and Schema Grounding)}, {\method} introduces \textbf{Semantic-Structural Distillation and
Stage-Aware Allocation} that integrates conceptual semantics, structured metadata abstractions, and syntax specifications into the generation process. Then the decoupled knowledge is elaborately distilled and processed. {\method} dynamically allocates them to the corresponding reasoning stages, ensuring that grounding information is incorporated rather than merely appended to the prompt. This design reduces semantic ambiguity and enables more reliable alignment.

For the challenge of Execution Reliability and Semantic Consistency (\textbf{C2}), {\method} introduces an \textbf{execution-guided revision loop and verification mechanism}. Critic Agent performs delta-guided memory management across iterations, explicitly identifying error evolution and repair effectiveness while maintaining a compact memory. Reviser Agent then leverages this feedback to perform targeted program repair and localized correction. Together, they form a tightly coupled feedback loop that progressively resolves execution errors. Beyond execution correctness, a final semantic verification stage explicitly validates consistency between the synthesized program and the original analytical intent, mitigating false positives that cannot be detected through execution feedback alone.

We evaluate {\method} on Spider 2.0-Lite and BIRD-SQL Mini-Dev. Experimental results demonstrate state-of-the-art performance across diverse backbone models while maintaining high token efficiency. On Spider 2.0-Lite, {\method} achieves up to \textbf{51.13\%} execution accuracy, substantially outperforming prior methods under the same backbone settings. On BIRD-SQL Mini-Dev, it reaches \textbf{49.40\%} execution accuracy. Moreover, {\method} requires only \textbf{11.2K--13.8K} tokens per task with three major backbones, substantially reducing inference cost while maintaining state-of-the-art accuracy. The framework generalizes effectively from lightweight open-source models (e.g., Qwen3-14B) to state-of-the-art commercial LLMs, and extensive ablation studies confirm the contribution of each major component.

In summary, our contributions are as follows:
\begin{itemize}

\item We propose {\method}, a framework for domain-constrained analytical program synthesis that unifies schema grounding, execution-guided refinement, and semantic verification to improve the reliability of LLM-powered analytics.

\item We develop Semantic-Structural Distillation and Stage-Aware Allocation for knowledge grounding, a mechanism that refines domain semantics and schema knowledge and allocates them to the corresponding synthesis stages.

\item We introduce an execution-guided refinement and semantic verification mechanism that performs delta-based failure diagnosis, targeted program repair, and intent-level validation to reduce executable-but-incorrect programs.

\item Extensive experiments on Spider 2.0-Lite and BIRD-SQL Mini-Dev demonstrate state-of-the-art execution accuracy, superior token efficiency, and strong generalization across both open-source and commercial LLM backbones.

\end{itemize}

\section{Related Works}
\label{section::related_works}


Recent LLM-powered analytics systems mainly differ in whether LLMs serve as intent translators that orchestrate external analytical engines or directly act as analytical engines themselves~\cite{wang2022interactive,maddigan2023chat2vis}.

\subsection{Code as Engine: Translators and Tool Users}

This paradigm represents the most mature and widely adopted approach, where LLMs function as intent translators that convert natural language requests into executable artifacts, typically SQL queries or Python programs, which are then executed by reliable external engines such as DBMSs or interpreters.
Its effectiveness is grounded in the strong code generation capabilities of modern LLMs~\cite{chen2021evaluating,lai2023ds}, and has been further extended by agent-based systems that enable tool use for multi-step analysis, including general platforms such as \textbf{OpenAgents}~\cite{xie2023openagents} and task-specific solutions in Text-to-SQL, e.g., divide-and-conquer parsing in \textbf{DIN-SQL}~\cite{pourreza2023din}, multi-agent collaboration in \textbf{CHESS}~\cite{talaei2024chess}, and prompt-level optimization in \textbf{DailSQL}~\cite{gao2023text}.

Despite their effectiveness, these approaches remain vulnerable to execution failures and executable-but-incorrect programs caused by incomplete grounding or flawed reasoning~\cite{wang2025towards,wang2024large}. Moreover, most systems employ loosely coupled retrieval, generation, and verification stages, limiting the use of execution feedback during synthesis. Recent methods such as \textbf{ReFoRCE}~\cite{deng2025reforce} and \textbf{AutoLink}~\cite{wang2025autolink} improve reliability through candidate exploration, but at the cost of substantial token and inference overhead.

\subsection{LLM as Engine: Direct Data Reasoner}

This line of work explores a more nascent and experimental paradigm, where LLMs are elevated to act as direct analytical engines, reasoning over data without relying on external execution environments.
Representative approaches follow two main directions: modular reasoning frameworks that decompose analytical tasks into predefined operators or reusable components~\cite{sartori2025combinatorial}, exemplified by systems such as \textbf{Galois}~\cite{saeed2023querying} and code-centric models that retrieve and adapt program fragments (e.g., \textbf{Codex}~\cite{chang2024systematic}, \textbf{Code Llama}~\cite{roziere2023code}); and table-specialized LLMs that aim to natively process and reason over tabular data, including \textbf{TableGPT}~\cite{zha2023tablegpt,li2024table}, \textbf{TableLlama}~\cite{zhang2024tablellama}, and \textbf{TableLLM}~\cite{zhang2025tablellm}.

Despite their promise, these approaches lack the execution guarantees provided by traditional analytical engines, making them difficult to scale to large or multi-relational datasets and more susceptible to semantic, factual, and arithmetic errors during complex analytical reasoning~\cite{zha2023tablegpt,zhang2024tablellama}.

\section{methodology}
\label{section::methodology}

\begin{figure*}[!t]
    \centering
    \includegraphics[width=\textwidth]{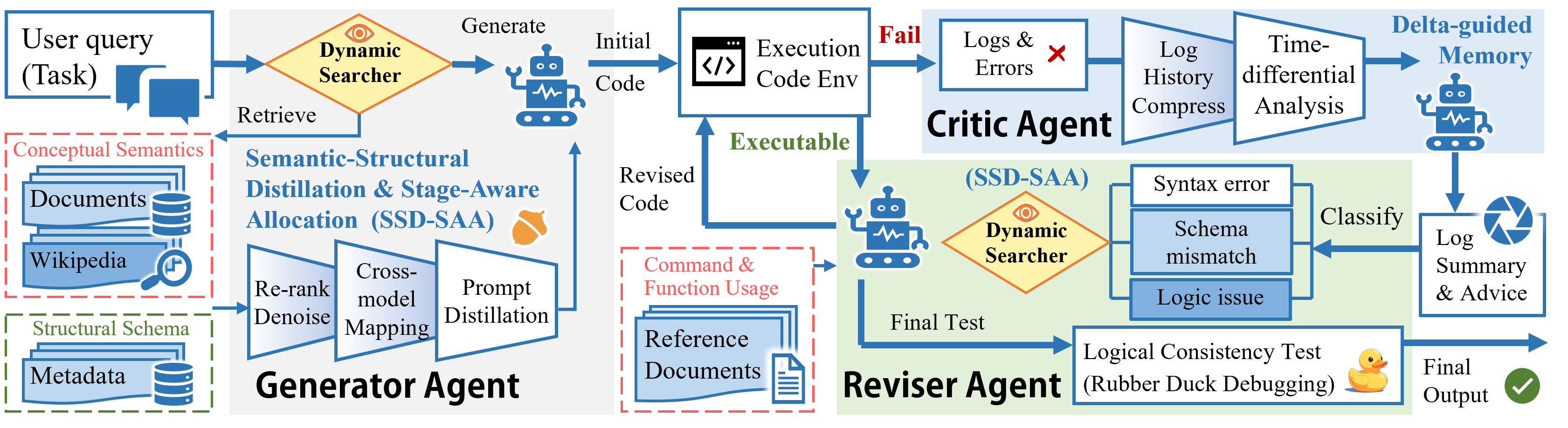}
    \caption{The Framework of the Multi-agent System}
    \label{fig::Framework}
\end{figure*}


The proposed framework, {\method} (Figure~\ref{fig::Framework}), formulates LLM-powered analytics as a \emph{domain-constrained analytical program synthesis} problem. Rather than relying on single-pass code generation, {\method} introduces a unified framework that jointly addresses two fundamental challenges: \emph{domain and schema grounding} and \emph{execution reliability with semantic consistency}. The framework consists of three tightly coordinated components: Generator, Critic, and Reviser.


To address the grounding challenge, {\method} introduces a \textbf{Semantic--Structural Distillation and Stage-Aware Allocation (SSD-SAA)} mechanism. A key observation is that retrieving relevant knowledge does not necessarily imply effective utilization during program synthesis. SSD-SAA separates grounding information into semantic and structural knowledge, refines them into compact synthesis-oriented representations, and dynamically allocates them to the corresponding reasoning stages. By explicitly coordinating knowledge usage with intent understanding, schema grounding, and code generation, this design establishes stronger alignment among analytical intent, database schemas, and executable program structures.

To ensure execution reliability and semantic consistency, {\method} further adopts the self-revision and verification loop. The Critic performs \textbf{delta-guided memory management} across iterations and maintains a compact history representation of historical debugging signals, allowing the framework to identify error evolution and repair effectiveness. Guided by this feedback, the Reviser performs targeted program repair for syntax, schema, and logical errors. A final logical verification stage then validates consistency between the synthesized program and the original analytical intent through intent-level reasoning, mitigating executable-but-incorrect failures that cannot be detected through execution feedback alone.

These components transform analytical program synthesis from a single-pass language modeling task into a closed-loop process of grounding, diagnosis, repair, and verification, which substantially improves execution correctness, semantic fidelity, and robustness in complex domain-specific analytical workloads.

\subsection{Generator Agent}

The Generator is responsible for transforming analytical intent into an initial executable program through explicit domain and schema grounding. To this end, we introduce \textbf{Semantic--Structural Distillation and Stage-Aware Allocation}, which decomposes grounding into three stages: \textbf{Decoupled Knowledge Acquisition}, \textbf{Knowledge Refinement}, and \textbf{Stage-Aware Allocation}. 
SSD-SAA converts heterogeneous knowledge sources into synthesis-oriented representations and dispatches them to the reasoning stages where they are most useful.

\subsubsection{Decoupled Knowledge Acquisition}

\begin{figure}[t]
    \centering
    \includegraphics[width=.5\textwidth]{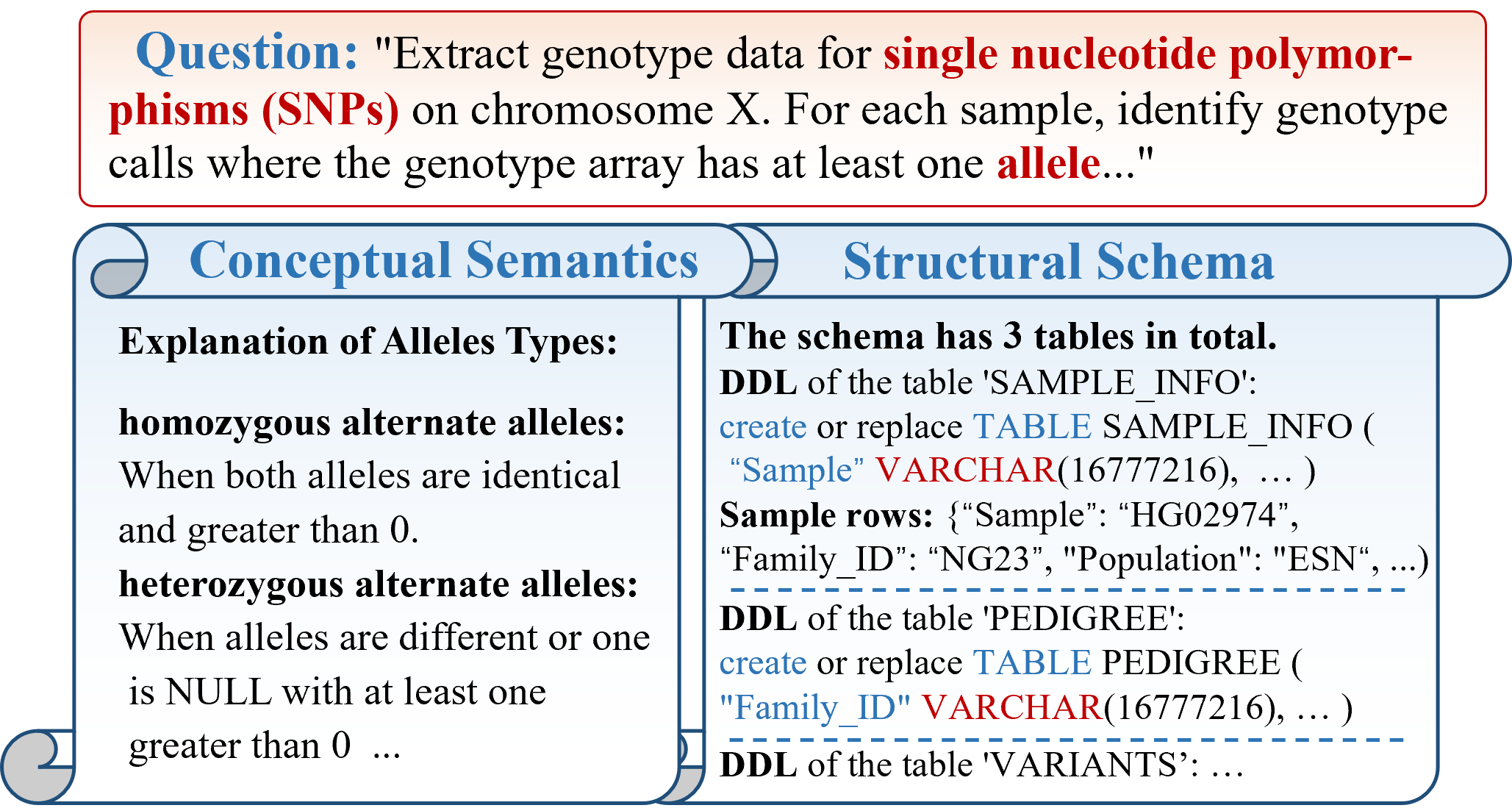}
    \caption{Decoupled Domain Knowledge Example}
    \label{fig::external_knowledge}
\end{figure}

{\method} adopts a \textbf{decoupled grounding strategy} that separates multi-source semantic knowledge into \textit{Conceptual Semantics} and \textit{Structural Schema} (Figure~\ref{fig::external_knowledge}) and processes them independently based on their own unique characteristics.

First, for \textit{Conceptual Semantics}, {\method} constructs a semantic grounding context through a hybrid strategy that combines dense knowledge in official documents with keyword-based matching, capturing both domain-specific identifiers and analytical concepts. To improve robustness under domain drift and previously unseen terminology, we further introduce a \textbf{Just-in-Time semantic expansion} mechanism that selectively supplements insufficient retrieval results using external knowledge sources (e.g., Wikipedia). Once the LLM makes the judgment that its parametric knowledge and document information are insufficient, it will activate the mechanism to acquire supplementary semantics, which are subsequently filtered to remove noisy content before being incorporated into the synthesis context. 

Then, for \textit{Structural Schema}, {\method} performs schema grounding through dedicated \textbf{Metadata Abstraction}. Unlike conventional approaches that directly serialize raw metadata into prompts, the module extracts and normalizes heterogeneous metadata sources—including table schemas, column descriptions, data types, DDL statements, and JSON specifications—into a unified canonical representation. To reduce context redundancy, homogeneous structures (e.g., partitioned tables sharing identical schemas) are compactly summarized while preserving execution-critical constraints and inter-table relationships. The abstraction provides an execution-oriented view of the underlying organization and serves as the structural grounding source for program synthesis.

\subsubsection{Knowledge Refinement}
After acquisition, the two knowledge sources remain crude and cannot be directly consumed for staged synthesis. To bridge the gap between semantic concepts and executable program structures, SSD-SAA applies a progressive knowledge refinement process that transforms raw results into a compact, stage-dispatchable representation. Concretely, this process performs: 
(1) \textbf{Relevance-aware re-ranking}: Queries, documents, and schema descriptions are encoded by BGE-M3 and ranked by cosine similarity to retain the top-$k$ evidence items.
(2) \textbf{Semantic--structural alignment}: Retrieved concepts are linked to corresponding schema elements and operator specifications.
(3) \textbf{Knowledge distillation}: The aligned knowledge is compressed into a compact synthesis-oriented representation that preserves essential constraints while minimizing context overhead.

\begin{figure}[t]
    \centering
    \includegraphics[width=.5\textwidth]{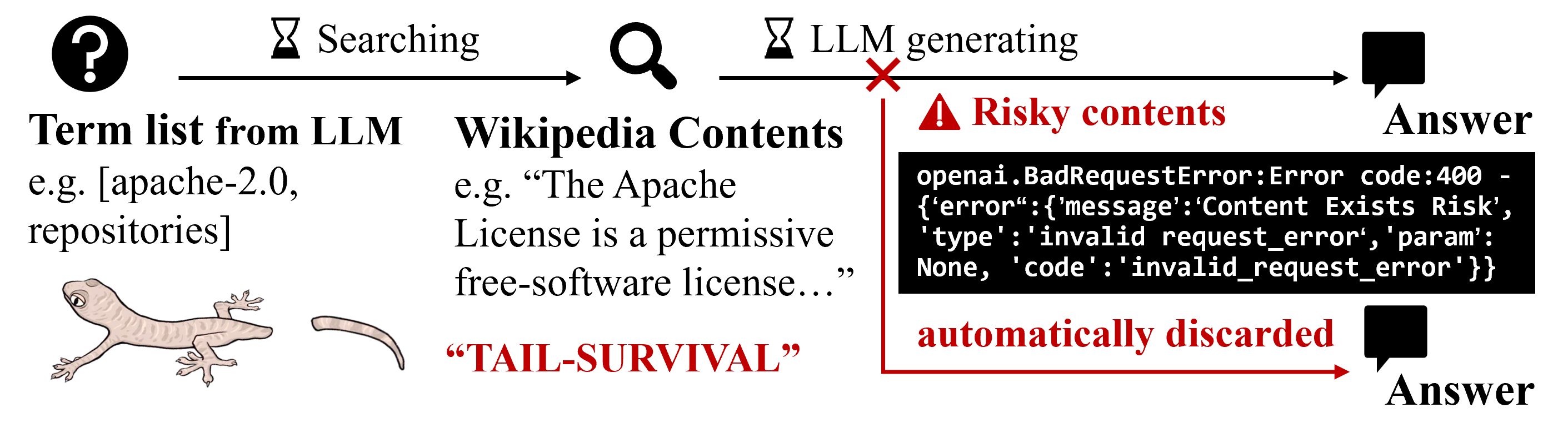}
    \caption{Risks Contents Fallback Mechanism}
    \label{fig::risks}
\end{figure}

We further integrate \textbf{fallback mechanism} to enhance robustness under real-world deployment conditions, as illustrated in Figure~\ref{fig::risks}. Unpredictable or noisy web-retrieved content may trigger internal safety filters and interrupt normal generation. When it happens, the system automatically discards risky external text while preserving essential schema and metadata context, allowing the Generator agent to continue functioning. This design considers the ranking and priority of heterogeneous semantics, which ensures graceful degradation rather than complete failure.


\subsubsection{Stage-Aware Knowledge Allocation}
Rather than injecting all refined knowledge at once, SSD-SAA orchestrates synthesis as a stage-aware allocation process that decomposes program generation into three tightly scoped sub-tasks: semantic intent analysis, schema grounding, and code formulation. During semantic intent analysis, only the distilled conceptual knowledge $K^*_{sem}$ is supplied to disambiguate analytical concepts and user objectives. During schema grounding, the distilled structural schemas $K^*_{struct}$ is incorporated to map the resolved intent to concrete database elements. Finally, during code formulation, the operator-level constraints and syntax specifications from official references $K^*_{syn}$ are applied to generate the executable analytical program. This explicit stage-conditioned allocation ensures that each category of knowledge is consumed precisely at the stage where it is most relevant, avoiding the inefficient utilization and weak grounding that plague undifferentiated knowledge injection. The resulting synthesis pipeline yields analytical programs with strong semantic fidelity and high execution reliability.

\definecolor{comment}{RGB}{0, 150, 0}
\begin{algorithm}[t]
\caption{SSD-SAA: Semantic-Structural Distillation and Stage-Aware Allocation (Generator Agent)}
\label{alg:ssdsaa}
\begin{algorithmic}[1]
\Require Query \textcolor{cyan}{$Q$}, Documents \textcolor{cyan}{$Doc$}, Wikipedia \textcolor{cyan}{$\mathcal{W}$}, Schema \textcolor{cyan}{$\mathcal{S}$}, Syntax and Function Usage \textcolor{cyan}{$Syn$}
\Ensure Initial Code Generation \textcolor{cyan}{$Code$}

\State \textcolor{gray}{\texttt{// 1. Decoupled Knowledge Acquisition}}
\State \textcolor{cyan}{$K_{sem}$} $\leftarrow$ \textsc{HybridRetrieve}$(\textcolor{cyan}{Q}, \textcolor{cyan}{Doc},\textcolor{cyan}{W})$
\Comment{\textcolor{comment}{Semantic concepts with JIT expansion}}
\State \textcolor{cyan}{$K_{struct}$} $\leftarrow$ \textsc{SchemaAbstract}$(\textcolor{cyan}{\mathcal{S}})$
\Comment{\textcolor{comment}{Normalized structural representation}}

\State \textcolor{gray}{\texttt{// 2. Knowledge Distillation \& Alignment}}
\State {\textcolor{cyan}{$K_{rele}$} $\leftarrow$ \textsc{Rerank}$(\textcolor{cyan}{Q}$, $K_{sem}$ $\cup$ {$K_{struct}$})}
\Comment{\textcolor{comment}{Relevance-aware filtering}}
\State \textcolor{cyan}{$K_{align}$} $\leftarrow$ \textsc{Align}({$K_{rel}^{sem}$}, {$K_{rel}^{struct}$})
\Comment{\textcolor{comment}{Explicit concept-to-schema mapping}}
\State \textcolor{cyan}{$K^*$} $\leftarrow$ \textsc{Distill}({$K_{align}$})
\Comment{\textcolor{comment}{Compress into high-signal compact form}}
\State \textcolor{cyan}{$(K_{sem}^*, K_{struct}^*)$} $\leftarrow$ \textsc{Partition}$(K^*)$
\Comment{\textcolor{comment}{Split knowledge by reasoning stage}}

\State \textcolor{gray}{\texttt{// 3. Stage-Aware Allocation}}
\State \textcolor{cyan}{$I$} $\leftarrow$ \textsc{IntentAnalysis}(\textcolor{cyan}{Q}, $K_{sem}^*$)

\State \textcolor{cyan}{$G$} $\leftarrow$ \textsc{SchemaGrounding}(\textcolor{cyan}{$I$}, $K_{struct}^*$)

\State \textcolor{cyan}{$Code$} $\leftarrow$ \textsc{Formulation}(\textcolor{cyan}{$G$}, $K_{op}^*$)

\State \Return \textcolor{cyan}{$Code$}
\end{algorithmic}
\end{algorithm}

\subsection{Critic Agent}
\label{section::critic}

Synthesizing executable and semantically faithful code remains a significant challenge for LLMs, particularly under complex analytical constraints. 
To address this issue, we design the Critic agent around a central principle: effective debugging requires not merely accumulating error logs, but maintaining \textbf{delta-guided memory management} that distills historical debugging experience and explicitly analyzes error transitions to inform subsequent revisions (Algorithm~\ref{alg:critic_agent}).

\subsubsection{Memory Adaptive Reflection}
A naive iterative strategy that repeatedly appends all historical errors and code versions quickly leads to excessive context growth, degrading efficiency and even exceeding the model's context window. 
To overcome this, the Critic implements this failure memory as an \textbf{adaptive reflection mechanism} that compactly summarizes past debugging data while preserving its diagnostic value.
Specifically, the agent maintains a rolling error history updated via a reflect-and-append mechanism: instead of retaining raw logs, it semantically compresses prior iterations $t-1$ and appends only the latest code–error pair:
$$H_{t} = \text{Reflect}(H_{t-1}) \oplus (\text{Code}_t + \text{Error}_{t})$$

This design introduces a principled \textbf{temporal decay}, analogous to momentum indicators in financial analysis~\cite{liu2021asymmetry}, using a time-weighted scheme to prioritize recent performance while gracefully mitigating stale information. Concretely, the summarized \textbf{long-term memory} captures recurring error patterns and their proven corrections, while the \textbf{short-term focus} remains anchored on the latest code and its corresponding execution error. This dual-granularity structure balances long-term pattern retention with short-term precision, eliminating redundancy while maintaining comprehensive awareness of the debugging trajectory.

\subsubsection{Delta-guided Failure Diagnosis}
Beyond memory compression, the Critic agent performs explicit \textbf{delta-guided failure diagnosis} by systematically comparing error transitions across successive iterations. By analyzing code and error deltas between iterations $t-1$ and $t$, the agent distinguishes two complementary diagnostic signals:
\begin{itemize}
\item \textbf{Constructive progression:} \(E_{t}\) differs in nature from \(E_{t-1}\). The previous correction (changes from \(Code_{t-1}\) to \(Code_{t}\)) successfully resolved the prior issue. This positive delta is recorded in debugging history as a validated solution pattern that offers reusable lessons for future debugging.
\item \textbf{Debugging stagnation:} \(E_{t}\) remains equivalent or closely related to \(E_{t-1}\). This negative delta signals that the current solution path is ineffective, and the agent is explicitly instructed to abandon it in favor of a divergent strategic alternative.
\end{itemize}

The core mechanism is the delta examination, which explicitly compares changes in code and error messages between consecutive iterations. This produces a meta-level diagnostic that prevents repetitive failures and actively steers the revision cycle away from proven dead ends toward more promising solution spaces. By grounding revision guidance in error patterns rather than isolated error instances, the Critic provides more actionable suggestions for the subsequent Reviser agent.

Overall, the Critic agent maintains a distilled, delta-aware reasoning trace that audits the debugging process, prevents logical drift, and stabilizes the refinement trajectory toward correctness. Operating under explicit execution supervision, it leverages both the current code state and blocker feedback to provide precise, transition-informed guidance to the Reviser agent.

\definecolor{comment}{RGB}{0, 150, 0}
\begin{algorithm}[t]
\caption{Delta-Guided Memory Management \\(Critic Agent)}
\label{alg:critic_agent}
\begin{algorithmic}[1]
\Require Current \textcolor{cyan}{$Code_t$}, Current Error \textcolor{cyan}{$E_t$}, Previous History \textcolor{cyan}{$H_{t-1}$}, Previous \textcolor{cyan}{$Code_{t-1}$}, Previous Error \textcolor{cyan}{$E_{t-1}$}
\Ensure Updated History \textcolor{cyan}{$H_t$}, Suggestion \textcolor{cyan}{$S_t$}

\State \textcolor{gray}{\texttt{// 1. Adaptive Reflection Mechanism (Temporal Decay)}}
\State \textcolor{cyan}{$H_{s}$} $ \leftarrow \text{Reflect}($ \textcolor{cyan}{$H_{t-1}$} $)$ \Comment{\textcolor{comment}{Distill long-term failure memory}}
\State \textcolor{cyan}{$H_t$} $\leftarrow$ \textcolor{cyan}{$ H_{s} $} $\oplus $ ( \textcolor{cyan}{$Code_t$}, \textcolor{cyan}{$E_t$} ) \Comment{\textcolor{comment}{Integrate current failure state}}

\State \textcolor{gray}{\texttt{// 2. Delta-Guided Failure Diagnosis}}
\State \textcolor{cyan}{$\Delta E$} $\leftarrow \text{Analyze}($ \textcolor{cyan}{$E_{t-1}$}, \textcolor{cyan}{$E_t$} )
\If{Error changes}\Comment{\textcolor{comment}{Positive delta: Fix works}}
    \State \textcolor{cyan}{$S_t$}$ \leftarrow \textbf{Constructive Progression}$ \Comment{\textcolor{comment}{Record as useful solution pattern in history}}
\Else \Comment{\textcolor{comment}{Negative delta: Fix failed}}
    \State \textcolor{cyan}{$S_t$}$ \leftarrow \textbf{Debugging Stagnation}$ \Comment{\textcolor{comment}{Explore divergent path}}
\EndIf
\State \Return \textcolor{cyan}{$H_t$}, \textcolor{cyan}{$S_t$}
\end{algorithmic}
\end{algorithm}

\subsection{Reviser Agent}
The Reviser is responsible for transforming partially correct programs into executable and semantically faithful programs. Unlike the Generator, which focuses on initial synthesis, the Reviser operates under execution supervision and leverages structured feedback from the Critic to perform targeted program repair and validation.

\subsubsection{Execution-Guided Program Repair}
The first responsibility of the Reviser is to correct execution failures. Given the delta-guided diagnosis generated by the Critic, the Reviser categorizes failures into three classes:

\begin{itemize}
\item \textbf{Syntax-level failures}, caused by invalid statements, incorrect function usage, or malformed expressions;
\item \textbf{Schema-level failures}, caused by incorrect table references, attribute mismatches, or invalid schema linking;
\item \textbf{Semantic-level failures}, caused by missing constraints, incorrect filtering conditions, or flawed analytical logic.
\end{itemize}

For each failure category, the Reviser selectively consults the corresponding grounding information provided by the same SSD-SAA module, including domain knowledge, schema metadata, and operator specifications. Rather than regenerating the entire program, the Reviser performs localized repair by identifying and modifying only the failure-inducing code segments. The revised program is then re-executed, forming an execution-guided refinement loop.

\subsubsection{Intent-Level Semantic Verification}

\begin{figure}[t]
    \centering
    \includegraphics[width=\linewidth]{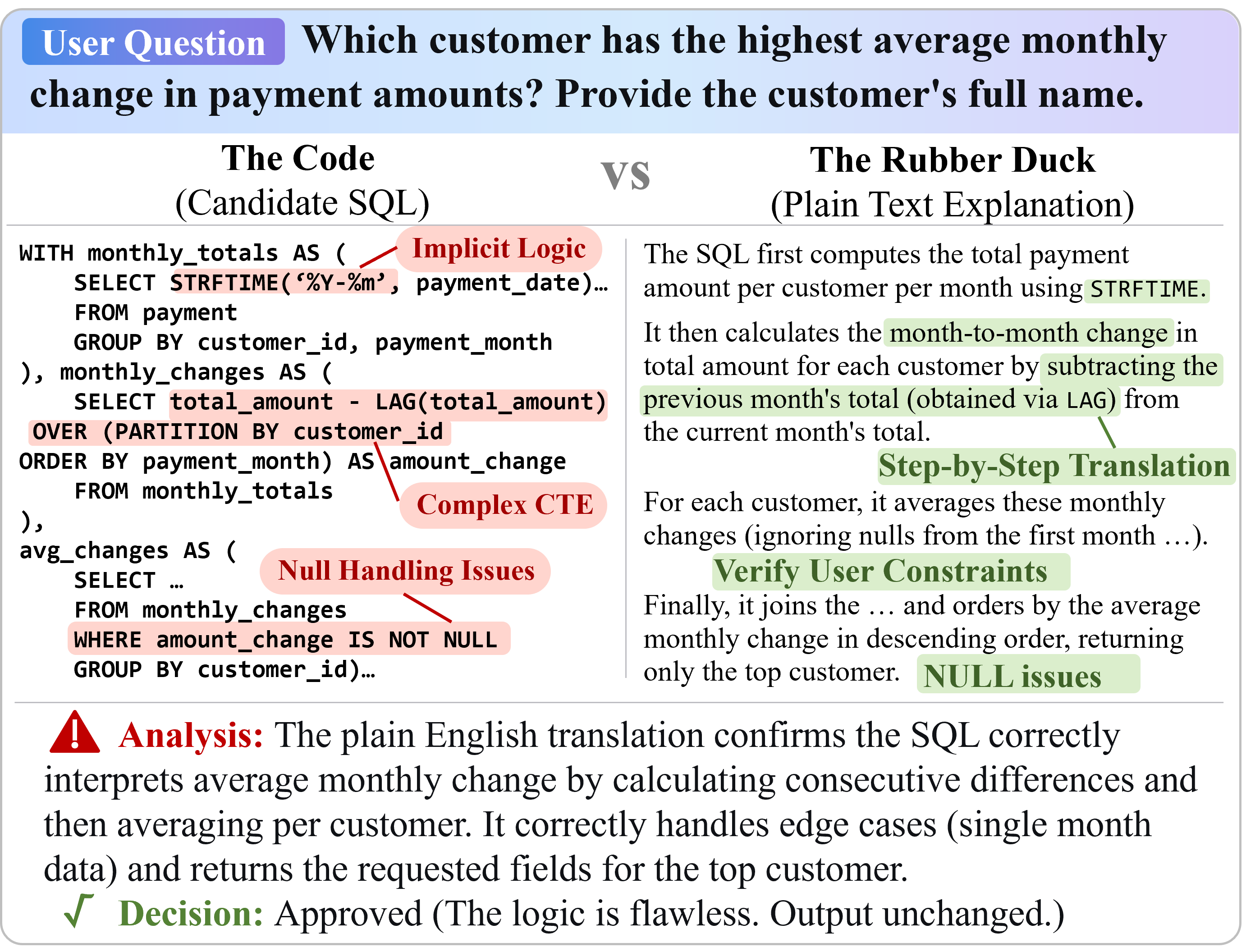}
    \caption{Rubber Duck Debugging Strategy}
    \label{fig:duck}
\end{figure}

Execution success does not necessarily imply analytical correctness. In practice, many generated programs satisfy execution constraints while violating the original analytical intent. To mitigate such executable-but-incorrect failures, the Reviser performs an additional semantic verification stage after execution succeeds.

A direct self-evaluation of generated code often suffers from confirmation bias, where the model tends to validate its own outputs~\cite{chen2023teaching, elvira2023digital}. To address this issue, we adopt a \textbf{Rubber Duck Debugging} strategy as an intent-level verification mechanism illustrated in Figure~\ref{fig:duck}. Specifically, the Reviser first translates the generated program into a structured natural-language explanation, explicitly exposing its analytical logic, including filtering conditions, aggregation operations, and output requirements. The explanation is then systematically compared against the original user request to verify semantic consistency.

When inconsistencies are detected, the Reviser triggers another round of targeted correction. Otherwise, the program is accepted as the final output. By combining execution-guided repair with intent-level semantic verification, the Reviser effectively reduces both execution failures and executable-but-incorrect programs.

\section{Experiment}
\label{sectionapi::experiment}

\begin{table*}[t]
\centering
\caption{Comparison of performance results on Spider 2.0-Lite}
\label{tab::spider_results}
\small
\begin{tabular}{@{}llccccc}
\toprule
\textbf{Model} & \textbf{Method} & \textbf{Easy} & \textbf{Medium} & \textbf{Hard} & \textbf{Total EX} & \textbf{Avg. Tokens} \\
\midrule
\multirow{3}{*}{GPT-4o}
    & DIN-SQL & 5.79 & 0.43 & 0.00 & 1.46 & — \\
    & CHESS  & 9.92 & 3.00 & 1.24 & 3.84 & — \\
    & DAIL-SQL  & 13.20 & 5.58 & 1.24 & 5.68 & — \\
    \rowcolor{gray!10}
     & \textbf{LYNX} (ours)     & 44.36 & 37.25  & 22.14 & \textbf{34.05} & 13.8K \\
\midrule
\multirow{2}{*}{DeepSeek-R1}
    & RSL-SQL       & —     & —     & —     & 30.53 & 50.0K \\
    & LinkAlign & — & — & — & 33.09 & — \\
    & AutoLink   & — & — & — & 34.92 & 38.0K \\
    \rowcolor{gray!10}
    & \textbf{LYNX} (ours)          & 68.00     &53.08     & 35.92     & \textbf{51.13} & \textbf{11.2K} \\
\midrule
\multirow{3}{*}{DeepSeek-V3}
& Spider-Agent & 20.31 & 6.10 & 4.05 & 8.78 & — \\
& LinkAlign & — & — & — & 24.86 & — \\
    & RSL-SQL       & —     & —     & —     & 26.14 & —   \\
    \rowcolor{gray!10}
    & \textbf{LYNX} (ours)          & 45.75$\pm$0.72 & 38.35$\pm$0.89 & 23.97$\pm$1.76 & \textbf{35.45}$\pm$1.02 & 13.4K \\
\bottomrule
\end{tabular}
\end{table*}

\subsection{Experimental Setup}
\label{subsection::experiment_setup}

\subsubsection{Testbed}

To ensure fair and interpretable comparisons, we adopt \textbf{GPT-4o}, \textbf{DeepSeek-V3}, and \textbf{DeepSeek-R1} as the primary backbones, as they are commonly used in recent NL-to-SQL and LLM-agent benchmarks. For all settings, the decoding temperature is fixed to \temperature\ to ensure reproducibility. To study backbone sensitivity and generalization, we further evaluate {\method} on a diverse set of models spanning both open-source and proprietary LLMs, including \textbf{Qwen3-14B}, \textbf{Claude-Sonnet-4.6}, and \textbf{Gemini-3.1-Pro}.


\subsubsection{Workloads} 

We evaluate {\method} on two widely used text-to-SQL benchmarks: \textbf{BIRD-SQL Mini-Dev}~\cite{li2024can} and \textbf{Spider 2.0-Lite}~\cite{lei2024spider}. BIRD-SQL Mini-Dev contains 500 instances derived from realistic databases and domain-specific analytical scenarios, while Spider 2.0-Lite contains 547 \footnote{Task sf009, sf013 and sf029 report database unavailable.} instances spanning BigQuery, Snowflake, and SQLite environments. Spider 2.0-Lite further emphasizes realistic and complex analytical workflows, with all SQL queries exceeding 50 whitespace-tokenized tokens and categorized into three difficulty levels based on token count: easy (<80 tokens), medium (80$\sim$159 tokens), and hard ($\geq$160 tokens). Together, these benchmarks evaluate both semantic grounding and execution reliability in real-world analytical tasks.



\subsubsection{Evaluation Metrics}

Following the official evaluation protocols of BIRD-SQL Mini-Dev and Spider 2.0-Lite~\cite{li2023can,lei2024spider}, we use \textbf{Execution Accuracy (EX)} as the primary metric, which measures whether the generated analytical program produces the correct execution result.

For BIRD-SQL Mini-Dev, we additionally report \textbf{Valid Efficiency Score (VES)} and \textbf{Soft-F1}. VES jointly evaluates execution correctness and efficiency, while Soft-F1 measures the semantic similarity between the predicted and ground-truth result sets at the cell level~\cite{li2023can}. Together, these metrics provide a comprehensive evaluation of analytical correctness, execution efficiency, and result-level semantic fidelity.

\subsubsection{Baselines}

We compare {\method} against representative methods spanning prompt-based, agent-based, and retrieval-enhanced text-to-SQL systems.
\textbf{Prompt-based methods} include \textbf{DIN-SQL}~\cite{pourreza2024din}, \textbf{DAIL-SQL}~\cite{gao2023text}, and \textbf{CHESS}~\cite{talaei2024chess}, which improve SQL generation through task decomposition, demonstration selection, and structured prompting.
\textbf{Agent-based systems} include \textbf{Spider-Agent}~\cite{lei2024spider}, which performs iterative query construction through tool interactions.
\textbf{Recent state-of-the-art systems} include \textbf{RSL-SQL}~\cite{cao2024rsl}, \textbf{ReFoRCE}~\cite{deng2025reforce}, \textbf{LinkAlign}~\cite{wang2025linkalign}, and \textbf{AutoLink}~\cite{wang2025autolink}, which incorporate schema grounding, retrieval enhancement, iterative refinement, and candidate exploration mechanisms to improve robustness on complex analytical workloads.


\subsection{Main Results}
\label{subsection::results}

\subsubsection{Overall Performance}


Table~\ref{tab::spider_results} reports the results on Spider~2.0-Lite. Following standard practice in recent NL-to-SQL literature, we report baseline numbers from their original papers or official releases. To ensure fairness, {\method} is evaluated using the same backbone models adopted by the corresponding baselines, allowing the comparison to focus on framework design rather than model capability. Overall, \textbf{LYNX consistently achieves the highest execution accuracy across all evaluated backbones while maintaining substantially lower token consumption}.
With GPT-4o, LYNX achieves an EX of \textbf{34.05\%}, significantly outperforming DAIL-SQL (5.68\%). Under DeepSeek-V3, LYNX reaches an average of \textbf{35.45\%} ($\pm$1.02), exceeding LinkAlign (24.86\%) and RSL-SQL (26.14\%), while the low variance across repeated runs demonstrates stable performance. When instantiated with DeepSeek-R1, LYNX achieves the highest EX of \textbf{51.13\%}, outperforming AutoLink (34.92\%) and RSL-SQL (30.53\%). 
The gains are particularly evident on complex queries. We attribute this improvement to the complementary effects of SSD-SAA and the delta-guided refinement loop. SSD-SAA strengthens domain and schema grounding during synthesis, while the Critic--Reviser feedback loop progressively eliminates execution and semantic errors, reducing the brittleness of single-pass generation and improving analytical correctness under complex reasoning workloads.


\subsubsection{Token Efficiency}

We measure token consumption as the cumulative prompt and completion tokens used per task until the framework produces an executable program that passes semantic verification. Since token usage grows with the number of generation and revision rounds, it serves as a practical indicator of inference cost.
As shown in Table~\ref{tab::spider_results}, {\method} requires only \textbf{11.2K--13.8K} tokens per task with three major backbones. In comparison, AutoLink and RSL-SQL consume \textbf{38.0K} and \textbf{50.0K} tokens, respectively. This substantial reduction in token consumption stems from our execution-guided repair strategy, which incrementally refines a single candidate program rather than relying on expensive candidate exploration, enabling {\method} to achieve both higher accuracy and lower inference cost.




\begin{table}[t]
\centering
\small
\caption{LYNX performance on different backbones}
\label{tab::models}
\begin{tabular}{lccc}
\toprule
\textbf{Backbone Model} & \textbf{EX \%} & \textbf{Avg. Times} & \textbf{Avg. Tokens} \\
\midrule
Qwen3-14B  & 10.66 & 6.50 & 48.4K \\
GPT-4o     & 34.05 & 5.16 & 13.8K \\
DeepSeek-V3 & 35.45 & 4.25 & 13.4K \\
DeepSeek-R1 & 51.13 & 4.80 & 11.2K \\
Claude-Sonnet-4.6 & 57.58 & 4.29  & 7.1K \\
Gemini-3.1-Pro & 59.54 & 3.70 &  6.4K \\
\bottomrule
\end{tabular}
\end{table}

\subsubsection{Generalization Across Models}

Table~\ref{tab::models} evaluates {\method} across a diverse set of backbone LLMs, ranging from lightweight open-source models to recent frontier models. Overall, LYNX generalizes well across a broad spectrum of model capabilities and consistently benefits from advances in stronger foundation models.
With frontier models such as \textbf{DeepSeek-R1}, \textbf{Claude-Sonnet-4.6}, and \textbf{Gemini-3.1-Pro}, LYNX consistently achieves more than \textbf{50\%} EX while requiring fewer refinement iterations, resulting in lower token consumption. This trend indicates that stronger backbones can more effectively leverage the grounding and execution-guided refinement mechanisms provided by LYNX.

\begin{figure}[t]
    \centering
    \includegraphics[width=\linewidth]{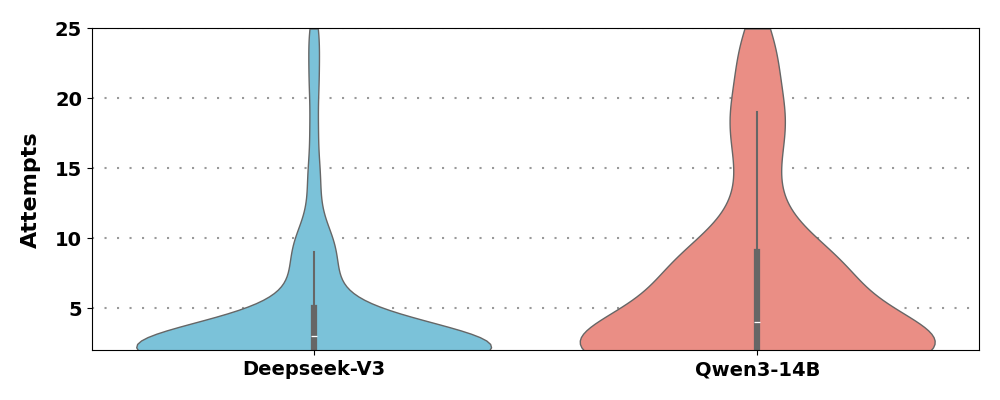}
    \caption{Attempt Distribution of Models}
    \label{fig:attempts}
\end{figure}

At the same time, the framework remains effective on lightweight models. As shown in Table~\ref{tab::models}, \textbf{Qwen3-14B} still achieves \textbf{10.66\%} EX, outperforming several existing systems built upon substantially stronger commercial backbones. This result suggests that the performance gains of LYNX are not solely derived from backbone capability, but also from its underlying synthesis architecture. Nevertheless, model capacity remains an important factor in determining the performance ceiling. Compared with Qwen3-14B, DeepSeek-V3 achieves substantially higher final executability (\textbf{88.42\%} vs. \textbf{49.36\%}) and requires fewer refinement iterations (\textbf{4.25} vs. \textbf{6.50}), indicating that stronger models can utilize execution feedback more effectively.
Figure~\ref{fig:attempts} further shows that a large proportion of Qwen3-14B's successful queries are obtained only after \textbf{6--10} refinement rounds. This observation suggests that execution-guided refinement is particularly valuable for lightweight models, where a considerable fraction of initially incorrect programs can still be repaired through iterative feedback.

\begin{table}[t]
    \centering
    \caption{Comparison of Executability Performances}
    \begin{tabular}{lcc}
    \toprule
    \textbf{Metric} & \textbf{DeepSeek-V3} & \textbf{Qwen3-14B} \\
    \midrule
    \textbf{Executability} & 88.42\% & 49.36\% \\
    \rowcolor{gray!15}
    \textbf{Avg. Attempts} & 4.25 & 6.50 \\
    \textbf{Few-Shot ($\leq$3 attempts)} & \multirow{2}{*}{68\%}  & \multirow{2}{*}{43\%} \\
    \textbf{Executable Rate} & & \\
    \bottomrule
    \end{tabular}
    \label{tab:model_comparison}
\end{table}

\subsubsection{Effect on BIRD-SQL Mini-Dev}

Table~\ref{tab:result_bird} reports the results on BIRD-SQL Mini-Dev, and {\method} consistently achieves the best performance across all evaluation metrics. With DeepSeek-V3, it attains \textbf{49.40\%} EX, \textbf{45.44\%} VES, and \textbf{56.31\%} Soft-F1, outperforming previous state-of-the-art methods. The simultaneous improvements in EX, VES, and Soft-F1 indicate that {\method} not only generates more correct analytical programs but also produces higher-quality execution results with better efficiency.
The gains are particularly notable when using lightweight models. When instantiated with \textbf{Qwen3-14B}, {\method} achieves \textbf{43.40\%} EX, surpassing TA~\cite{roziere2023code,qu2024before} built upon the substantially larger Llama3-70B backbone. This result suggests that the effectiveness of {\method} is not solely dependent on model scale, but also benefits from its grounding and execution-guided refinement mechanisms.
Compared with Spider 2.0-Lite, BIRD places greater emphasis on domain-specific semantics and realistic analytical scenarios. 
The results suggest that accurate grounding via SSD-SAA and iterative refinement via the Critic--Reviser loop jointly drive the performance gains. A detailed breakdown across difficulty levels is provided in Table~\ref{tab:BIRD_difficulty}.


\begin{table}[t]
    \centering
    \caption{Performance Comparison on BIRD Mini-Dev}    
    \begin{tabular}{lccc}
    \toprule
    \textbf{Methods (Models)} & \textbf{EX (\%)} & VES & Soft-F1 \\
    \midrule
    \textbf{TA} + codellama-34B~\cite{roziere2023code} & 27.57 & - & - \\
    GPT-3.5-turbo~\cite{ouyang2022training} & 38.00 & 37.33 & 41.84\\
    llama3-70B-instruct & 40.80 & 39.02 & 44.38\\
    Claude-2~\cite{anthropic2023introducing} & 42.70 & - & -\\
    \textbf{TA} + llama3-70B-instruct~\cite{qu2024before} & 42.80 & 41.37 & 46.66\\
    \textbf{TA} + DeepSeek~\cite{qu2024before} & 43.74 & - & - \\
    \textbf{Struct-SQL}~\cite{thaker2025knowledge} &  45.00 & - & -\\
    \midrule
    \textbf{{\method}(ours)}  + Qwen3-\textbf{14B} & \underline{43.40} & 40.48 & 49.92 \\
    \textbf{{\method}(ours)}  + Deepseek-V3 & \textbf{49.40} & \textbf{45.44} & \textbf{56.31}\\
    \bottomrule
    \end{tabular}
    \label{tab:result_bird}
\end{table}

\begin{table}[t]
\centering
\caption{Detailed Results across Difficulty Levels}
\begin{tabular}{p{1.6cm}p{0.9cm}>{\centering\arraybackslash}p{0.7cm}>{\centering\arraybackslash}p{1cm}cc}
\toprule
\textbf{Model} & \textbf{Metric} & \textbf{Simple} & \textbf{Moderate} & \textbf{Hard} & \textbf{Total} \\
\midrule
\rowcolor{gray!15}
Qwen3-14B & EX & 57.43 & 39.20 & 33.33 & 43.40 \\
          & VES & 54.11 & 36.10 & 31.44 & 40.48 \\
          & Soft-F1 & 62.95 & 45.22 & 42.55 & 49.92 \\
\midrule
\rowcolor{gray!15}
DeepSeek-V3 & EX & 63.51 & 47.20 & 34.31 & 49.40 \\
              & VES & 58.69 & 43.02 & 32.14 & 45.44 \\
              & Soft-F1 & 68.22 & 54.91 & 42.50 & 56.31 \\
\bottomrule
\end{tabular}
\label{tab:BIRD_difficulty}
\end{table}


\begin{figure}[t]
    \centering
    \includegraphics[width=\linewidth]{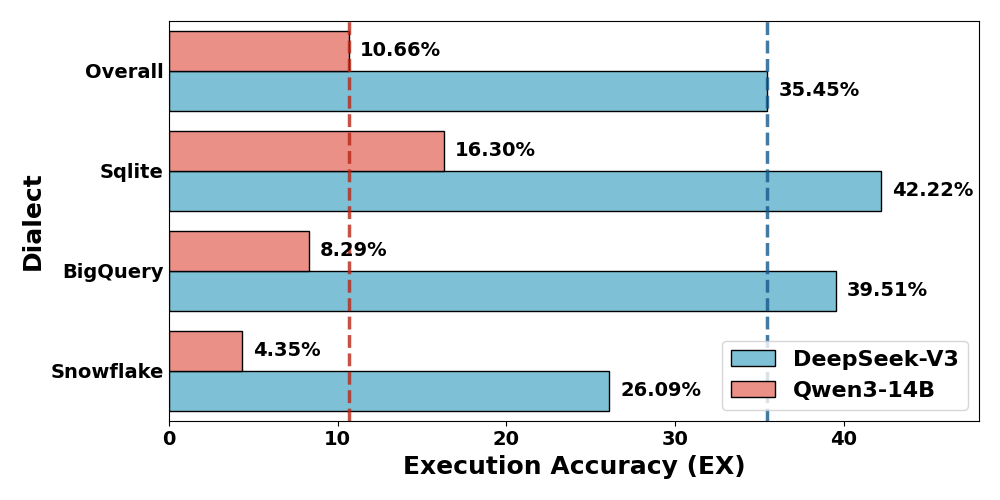}
    \caption{Scores on Different SQL Dialects}
    \label{fig:dialect}
\end{figure}

\subsubsection{Dialect Adaptability}

To understand the impact of dialect-specific syntax and schema characteristics, we further evaluate {\method} across multiple SQL dialects. Figure~\ref{fig:dialect} presents the corresponding execution accuracy of \textbf{SQLite, BigQuery, Snowflake}.
SQLite achieves the highest accuracy due to its relatively standardized syntax and simpler schema organization. In contrast, performance degrades on BigQuery and Snowflake, which introduce dialect-specific operators, proprietary functions, and more complex schema abstractions. The gap is particularly pronounced on Snowflake, where analytical programs often involve hierarchical object references (e.g., \texttt{DATABASE.SCHEMA.TABLE}), specialized semi-structured data operators, and dialect-specific function semantics. These characteristics increase the difficulty of both schema grounding and syntax generation, making execution correctness more challenging to achieve.
Despite these challenges, {\method} maintains consistent performance advantages across all dialects. The remaining performance gap indicates that dialect-specific reasoning remains a challenging problem for current LLMs and represents a promising direction for future improvement.

\definecolor{crimson}{RGB}{220,20,60}
\begin{table}[t]
\centering
\small
\caption{Ablation Results on Spider 2.0-lite}
\label{tab::ablation}
\begin{tabular}{llcc}
\toprule
\textbf{agent} & \textbf{function} & \textbf{EX \%} ($\Delta$)\\
\midrule
- & Full model (Ours) & 35.45 \\
\midrule
\textbf{\textcolor{crimson}{Generator}} &w/o SSD-SAA & $\approx 0$ (\textcolor{crimson}{$\downarrow \; \approx$100\%}) \\
\midrule
\textbf{\textcolor{crimson}{Critic}} &
w/o Delta-guided memory & 27.36 (\textcolor{crimson}{$\downarrow \;$22.8\%}) \\
\midrule
\multirow{2}{*}{\textbf{\textcolor{crimson}{Reviser}}} 
    & w/o Iterative revision & 22.43 (\textcolor{crimson}{$\downarrow \; $36.7\%}) \\
    & w/o Logical verification & 30.23 (\textcolor{crimson}{$\downarrow \; $14.7\%}) \\
\bottomrule
\end{tabular}
\end{table}

\subsection{Ablation Analysis}
\label{section::ablation}

\subsubsection{Agent Ablation}

Table~\ref{tab::ablation} reports the ablation results using DeepSeek-V3. Since the Generator, Critic, and Reviser form a tightly coupled synthesis pipeline, we perform ablation at the \textbf{functional level} rather than removing entire agents. Overall, all components contribute substantially to the final performance, confirming the effectiveness of the proposed closed-loop design.

\noindent
\textbf{Effect of Generator.} Removing SSD-SAA causes execution accuracy to collapse to nearly zero. Without explicit semantic and schema grounding, the model cannot reliably resolve domain terminology, schema elements, or structural constraints, making analytical program synthesis largely infeasible. This result highlights that accurate grounding is a prerequisite for complex NL-to-analytics tasks.

\noindent
\textbf{Effect of Critic.} Removing the Critic reduces EX from \textbf{35.45\%} to \textbf{27.36\%}. Without differential error analysis and reflection memory, the framework loses visibility into error evolution across iterations, making it difficult to distinguish effective repairs from repeated failures. This demonstrates the importance of execution-aware diagnosis for guiding iterative refinement.

\noindent
\textbf{Effect of Reviser.} Disabling execution-guided repair decreases EX from \textbf{35.45\%} to \textbf{22.43\%}, indicating that a substantial portion of generated programs require targeted correction before becoming executable. Furthermore, removing semantic verification reduces EX to \textbf{30.23\%}, showing that execution success alone does not guarantee analytical correctness. The additional improvement confirms the effectiveness of intent-level verification in mitigating executable-but-incorrect programs.

Overall, the results validate the complementary roles of the three components: Generator provides reliable grounding, the Critic supplies execution-aware diagnostic signals, and the Reviser progressively improves both execution correctness and semantic fidelity through iterative refinement.



\subsubsection{Overhead Analysis}

\begin{figure}[t]
    \centering
    \includegraphics[width=\linewidth]{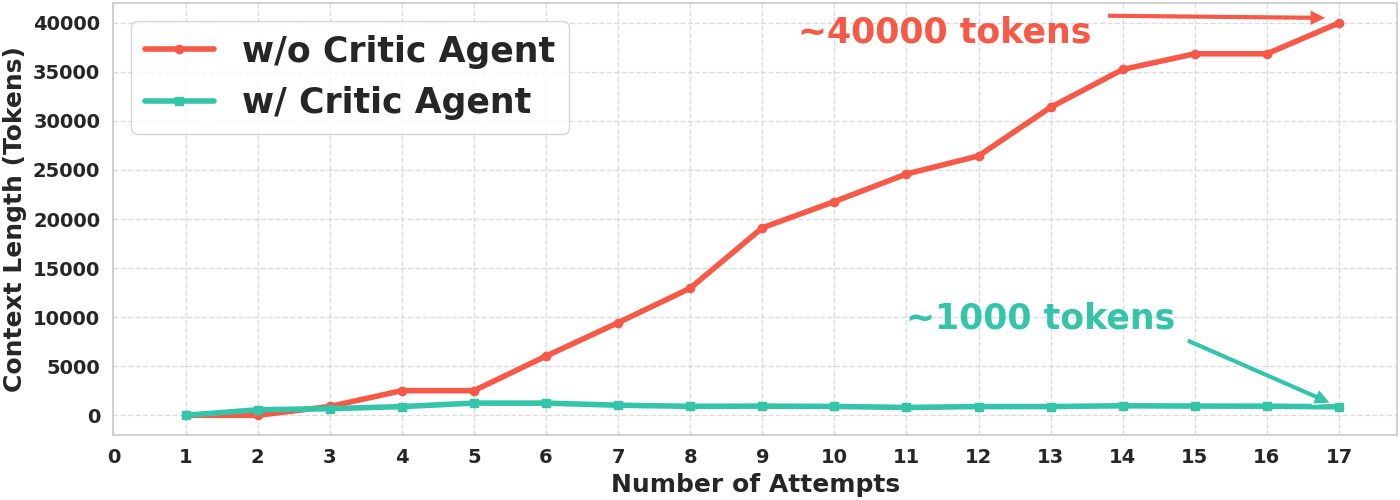}
    \caption{Comparison of History Length after Attempts}
    \label{fig:summary}
\end{figure}

Table~\ref{tab::token_consumption} reports the token consumption and runtime breakdown of the three components. The overhead remains moderate and aligns with the functional responsibilities of each component. The \textbf{Generator} incurs the highest runtime (14.94 $s/query$) and prompt cost, as it performs knowledge acquisition, refinement, and stage-aware grounding before synthesis. The \textbf{Critic} exhibits the largest completion length because it maintains compressed error traces and produces diagnostic feedback for subsequent repair. In contrast, the \textbf{Reviser} introduces the lowest prompt overhead, as it operates on structured diagnostic signals rather than raw execution contexts.
A key concern in iterative program synthesis is the growth of historical debugging context. To evaluate the effectiveness of the Critic's memory management mechanism, we track the length of \texttt{error\_history} across revision iterations (Figure~\ref{fig:summary}). Without compression, the accumulated debugging context grows approximately linearly and can exceed \textbf{40K} tokens, rapidly becoming the dominant source of inference cost. In contrast, the Critic maintains a compact representation through differential summarization, effectively bounding context growth while preserving diagnostic information.

\begin{table}[t]
\centering
\small
\caption{Token Consumption and Runtime Analysis}
\label{tab::token_consumption}
\begin{tabular}{lccc}
\toprule
\multirow{2}{*}{\textbf{Agent}} & \multicolumn{2}{c}{\textbf{Tokens}} & \multirow{2}{*}{\textbf{Runtime}} \\
\cmidrule(lr){2-3}
& \textbf{Prompt} & \textbf{Completion} & \\ 
\midrule
Generator  & $3520 \pm 627$ & $496 \pm 102$ & $14.94 \pm 4.08s $  \\ 
Critic     & $3462 \pm 1128$ & $934 \pm 228$ & $6.82 \pm 1.40s $ \\ 
Reviser   & $2438 \pm 413$ & $473 \pm 113$ & $ 10.16 \pm 3.92s $ \\
\bottomrule
\end{tabular}
\end{table}

\subsubsection{Stage-aware Reasoning Efficacy}

\begin{figure}[t]
    \centering
    \includegraphics[width=\linewidth]{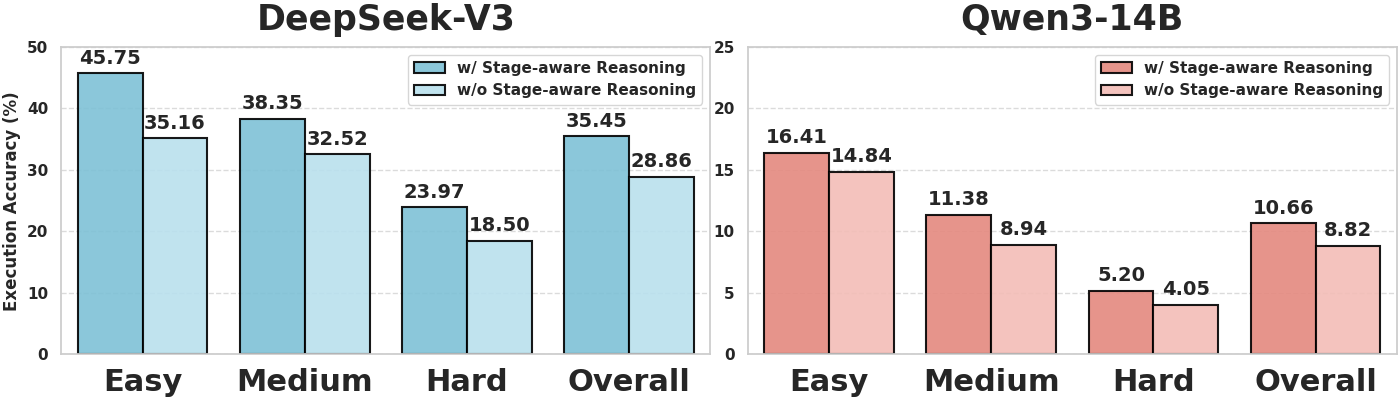}
    \caption{Details for Stage-aware Reasoning Efficacy}
    \label{fig:stage-aware-reasoning}
\end{figure}

Stage-aware reasoning is a core design principle of {\method} and is employed throughout the Generator, Critic, and Reviser. To evaluate its contribution, we remove all task decomposition and stage-specific reasoning instructions. As shown in Figure~\ref{fig:stage-aware-reasoning}, stage-aware reasoning consistently improves performance across different backbones and query difficulty levels, indicating that it provides a general benefit rather than targeting a specific query category. The improvement is particularly evident on DeepSeek-V3 (\textbf{+6.59\%} EX), compared with Qwen3-14B (\textbf{+1.84\%} EX), suggesting that stronger models are better able to exploit stage-aware reasoning and grounding signals during analytical program synthesis.


\subsubsection{Iterative Revision Loop}

To quantify the impact of execution-guided refinement, Table~\ref{tab:ablation_revision} compares the performance with and without the iterative revision loop. The results demonstrate substantial gains across models of different scales. For Qwen3-14B, execution accuracy more than doubles from \textbf{4.96\%} to \textbf{10.66\%}. Under DeepSeek-V3, EX increases from \textbf{22.43\%} to \textbf{35.45\%}. These improvements indicate that a significant fraction of initially generated programs can be repaired through iterative execution feedback. Notably, the relative gains are larger for Qwen3-14B, suggesting that execution-guided refinement is particularly valuable for lightweight models, where it compensates for weaker schema understanding and domain knowledge through progressive self-correction.


\subsubsection{BIRD-SQL Complementary Results}

Tables~\ref{tab:bird_ablation} reports the impact of semantic grounding on BIRD-SQL Mini-Dev. Compared with Spider 2.0-Lite, BIRD adopts relatively homogeneous SQLite schemas, making many schema structures and operators easier to infer from the model's parametric knowledge. Consequently, semantic grounding yields smaller gains in execution accuracy. Nevertheless, {\method} consistently improves result quality, increasing \textbf{Soft-F1} by \textbf{2.32} (specifically \textbf{+2.00} for hard subset). This suggests that semantic grounding remains beneficial even when execution accuracy saturates, as it improves semantic alignment between generated programs and analytical intent. The results further indicate that the effectiveness of grounding mechanisms becomes more pronounced as schema complexity and domain ambiguity increase.


\definecolor{gaincolor}{RGB}{0, 150, 0}

\begin{table}[t]
    \centering
    \caption{Revision Loop Effects across Models}
    \label{tab:ablation_revision}
    \begin{tabular}{>{\centering\arraybackslash}p{1.9cm}cc}
        \toprule
        \multirow{2}{*}{
            \diagbox[width=1.9cm, height=0.8cm]
                {\textbf{Setting}}
                {\textbf{Model}}
        }
        & \textbf{Qwen3-14B}
        & \textbf{DeepSeek-V3} \\
        \cmidrule(lr){2-2} \cmidrule(lr){3-3}
        & \textbf{EX (\%)}
        & \textbf{EX (\%)} \\
        \midrule
        w/o Revision & 4.96  & 22.43 \\
        w/ Revision  & 10.66 & 35.45 \\
        $\mathbf{\Delta}$
        & \textcolor{gaincolor}{5.70 ($\uparrow$115\%)}
        & \textcolor{gaincolor}{13.02 ($\uparrow$58\%)} \\
        \bottomrule
    \end{tabular}
\end{table}

\newcommand{\inc}[1]{\textcolor{gaincolor}{+#1}} 
\newcommand{\dec}[1]{\textcolor{crimson}{-#1}}  

\begin{table}[t]
\centering
\caption{Semantic Impact to Subsets on BIRD}
\label{tab:bird_ablation}
\begin{tabular}{llcccc}
\toprule
\multirow{2}{*}{\textbf{Metric}} & \multirow{2}{*}{\textbf{Method}} & \multicolumn{4}{c}{\textbf{Difficulty}} \\
\cmidrule(lr){3-6}
 & & Simple & Moderate & Hard & \textbf{Total} \\
\midrule
\multirow{3}{*}{\textbf{EX}} 
 & w/o Semantics  & 62.16 & 44.00 & \textbf{36.27} & 47.80 \\
 & w/ Semantics   & \textbf{63.51} & \textbf{47.20} & 34.31 & \textbf{49.40} \\
 & $\Delta$ & \inc{1.35} & \inc{3.20} & \dec{1.96} & \inc{1.60} \\
\midrule
\multirow{3}{*}{\textbf{VES}} 
 & w/o Semantics  & 56.74 & 40.17 & \textbf{34.23} & 43.86 \\
 & w/ Semantics  & \textbf{58.69} & \textbf{43.02} & 32.14 & \textbf{45.44} \\
 & $\Delta$ & \inc{1.95} & \inc{2.85} & \dec{2.09} & \inc{1.58} \\
\midrule
\multirow{3}{*}{\textbf{Soft-F1}} 
 & w/o Semantics & 65.83 & 52.48 & 40.50 & 53.99 \\
 & w/ Semantics  & \textbf{68.22} & \textbf{54.91} & \textbf{42.50} & \textbf{56.31} \\
 & $\Delta$ & \inc{2.39} & \inc{2.43} & \inc{2.00} & \inc{2.32} \\
\bottomrule
\end{tabular}%
\end{table}



\section{Case Study}

While aggregate metrics quantify overall performance, they provide limited visibility into how individual components contribute to analytical program synthesis. We present two representative case studies to examine the behavior of {\method} under realistic analytical workloads.
The first case study focuses on a quantitative finance task and illustrates how the Critic--Reviser loop identifies execution failures, tracks error evolution, and progressively repairs analytical programs. The second case study is drawn from Spider 2.0-Lite and demonstrates how SSD-SAA grounds domain semantics and schema information to support complex multi-table reasoning.


\subsection{Quantitative Finance Workloads}

Quantitative finance presents a challenging setting for analytical program synthesis due to its specialized terminology, complex operator semantics, and multi-step numerical reasoning. Many financial concepts (e.g., \textit{ADV}, \textit{VWAP}, and industry-neutralized factors) are underrepresented in the pretraining corpora of general-purpose LLMs, while factor computation often involves rolling-window statistics, cross-sectional ranking, and nested arithmetic operations.

\begin{figure}[t]
    \centering
    \includegraphics[width=\linewidth]{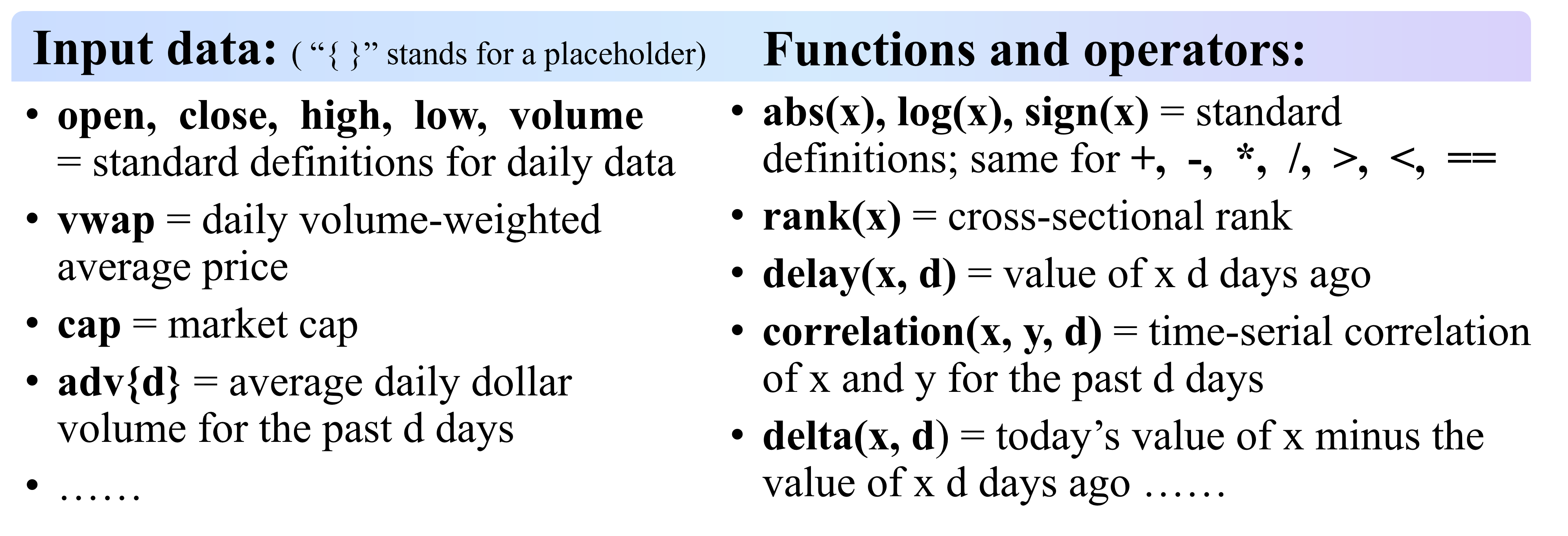}
    \caption{Definitions for External Knowledge}
    \label{fig:definitions}
\end{figure}

To evaluate whether {\method} generalizes beyond text-to-SQL benchmarks, we apply it to the computation of Alpha\#36 from the widely used \textbf{WorldQuant 101 Alphas}~\cite{kakushadze2016101}. The workload consists of CSV data for \textbf{100} stocks, each containing more than \textbf{3,900} daily observations and 33 financial attributes. In addition to raw data, formal definitions of financial indicators and domain-specific operators are provided as external knowledge (Figure~\ref{fig:definitions}).

\begin{figure*}[t]
    \centering
    \includegraphics[width=\linewidth]{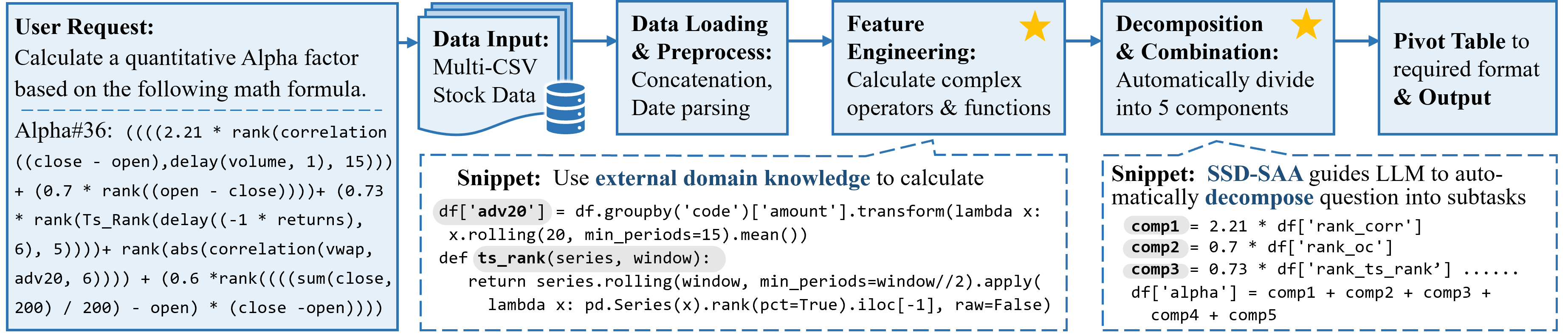}
    \caption{The Pipeline of the Generated Python Script}
    \label{fig:pipeline}
\end{figure*}

\begin{figure}[t]
    \centering
    \begin{minipage}[b]{0.4\linewidth}
        \centering
        \includegraphics[width=\linewidth]{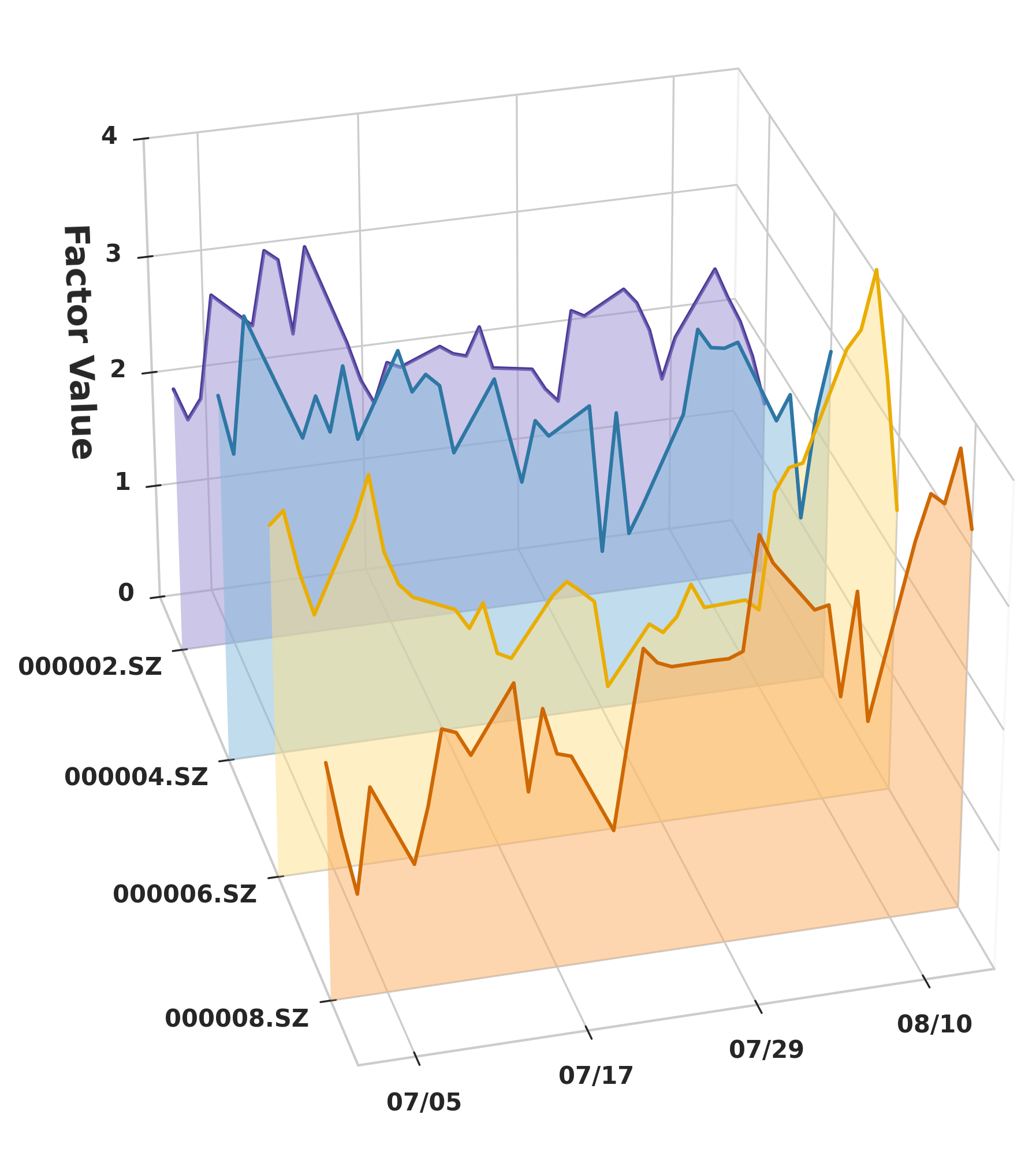}
    \captionof{figure}{Factor Results}
    \label{fig:factor_visual}
    \end{minipage}
    \begin{minipage}[b]{0.55\linewidth}
        \centering
    \includegraphics[width=\linewidth]{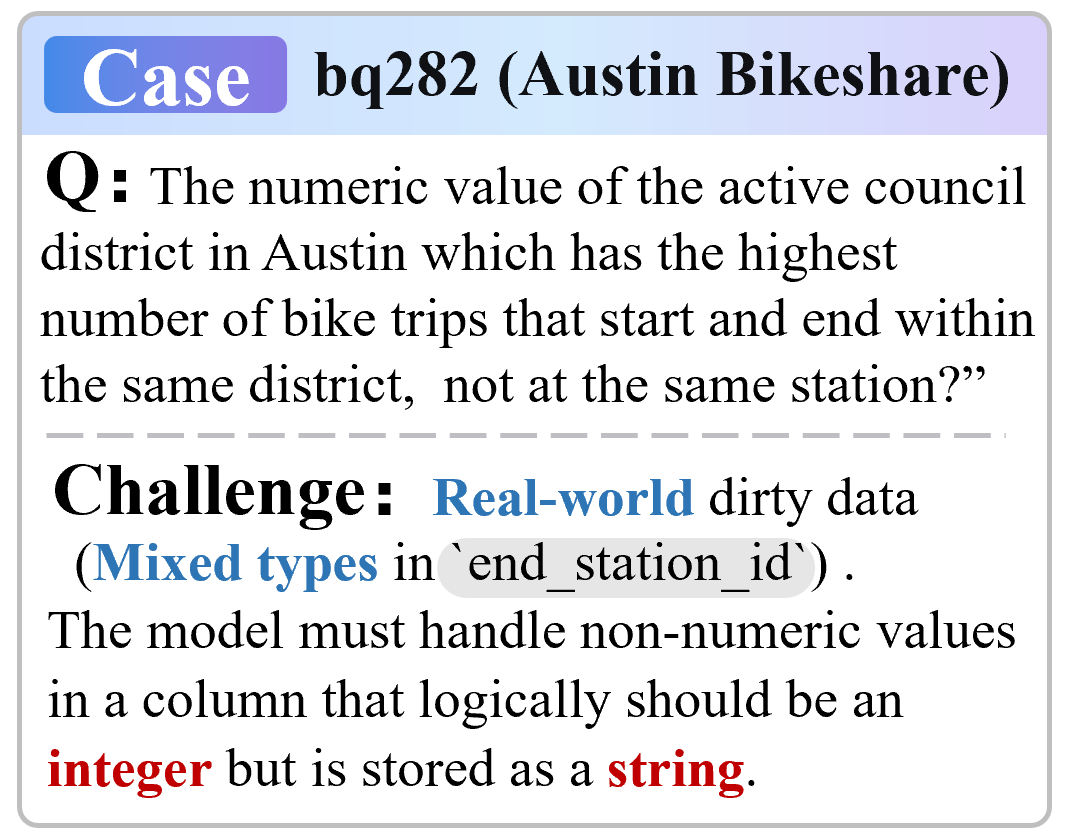}
    \captionof{figure}{Case Overview}
    \label{fig:overview}
    \end{minipage}
\end{figure}

As illustrated in Figure~\ref{fig:pipeline}, SSD-SAA first grounds financial terminology and operator semantics using the supplied domain knowledge. The Generator then synthesizes a multi-stage Python workflow covering data loading, preprocessing, intermediate feature computation, and factor aggregation, while the Critic--Reviser loop iteratively resolves execution and logical errors during synthesis. The resulting pipeline successfully computes Alpha\#36 over the complete dataset. Figure~\ref{fig:factor_visual} further shows the resulting factor distribution across stocks and time periods, indicating that the generated program produces meaningful and non-trivial financial signals rather than degenerate outputs. This demonstrates that {\method} can effectively combine domain grounding and iterative refinement to support complex numerical analytics beyond SQL generation.

\subsection{Spider 2.0-Lite Case}
\label{section::spider_case}

To complement the aggregate benchmark results, we examine a representative Spider 2.0-Lite instance (\textit{bq282}) that highlights the challenges of real-world analytical program synthesis. The task involves heterogeneous data containing mixed types and requires strict adherence to BigQuery-specific conventions. Even when powered by a frontier model (\textbf{GPT-5.2}), the generated query fails to execute due to incorrect handling of BigQuery \textbf{Fully Qualified Names (FQN)}, resulting in invalid table references. Furthermore, the query omits the required \textbf{safety casting} operations, making it vulnerable to malformed or missing values commonly encountered in real-world datasets.

\begin{figure}[t]
    \centering
    \includegraphics[width=\linewidth]{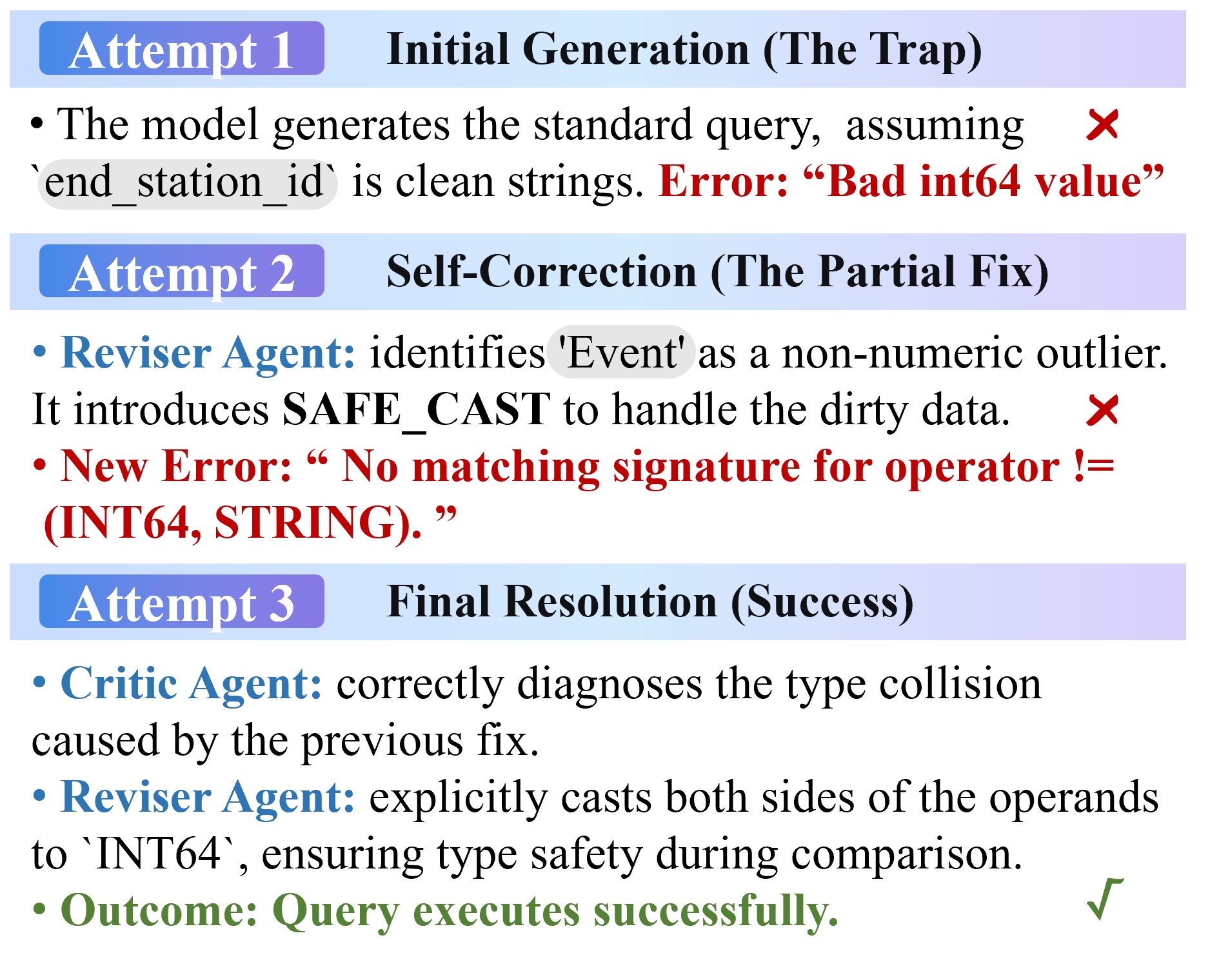}
    \vspace{-1em}
    \caption{Case Multi-step Process}
    \label{fig:process}
\end{figure}

Figure~\ref{fig:process} illustrates how {\method} resolves this failure through execution-guided refinement. Instead of treating execution errors as isolated events, the Critic performs differential analysis over successive error states and identifies the underlying causes of failure. Guided by these diagnostic signals, the Reviser incrementally repairs the query by correcting dialect-specific table references and introducing appropriate type-safe transformations. Through multiple refinement iterations, the generated program evolves from a non-executable query into a valid solution that satisfies both execution and semantic requirements.

A side-by-side comparison between the failed GPT-5.2 output and the final query generated by {\method} is shown in Table~\ref{tab:case_study_comparison}. This example demonstrates that strong language modeling capability alone is insufficient for reliable analytical program synthesis in realistic database environments. Instead, robust performance requires execution-aware diagnosis and iterative repair, which are explicitly supported by the Critic--Reviser feedback loop. The case therefore provides qualitative evidence for the effectiveness of the proposed execution-guided refinement mechanism beyond what can be observed from aggregate benchmark metrics alone.



\begin{table}[t]
    \centering
    \caption{Comparative Analysis of Case "bq282"}
    \footnotesize
    \begin{tabular}{lp{2.7cm}p{3cm}}
    \toprule
    \textbf{Stage} & \textbf{general-purpose LLM (GPT-5.2)} & \textbf{Our Framework}\newline \textbf{(w/ revision)} \\
    \midrule
    \textbf{Schema} & \textbf{Hallucinate} table name & use correct FQN \\
    \midrule
    \textbf{Challenge} & Standard casting fails on dirty data & Detected error and applied \texttt{SAFE\_CAST} \\
    \midrule
    \textbf{Logic} & Implicit type mismatch \newline in comparisons & Explicitly compare \newline \texttt{INT64} vs \texttt{STRING}\\
    \midrule
    \textbf{Outcome} & \textbf{\textcolor{red}{Execution Error}} \newline "404 Not Found" \quad \xmark & \textbf{\textcolor{olive}{Successfully executed} }\newline Correct Answer \hfill \cmark\\
    \bottomrule
    \end{tabular}
    \label{tab:case_study_comparison}
\end{table}






\section{Conclusion}

In this paper, we present {\method}, a multi-agent framework that formulates reliable LLM-powered analytics as a domain-constrained analytical program synthesis process. {\method} addresses two central challenges in analytical program generation: grounding analytical intent in domain semantics and database structure, and reliably refining generated programs toward both execution correctness and semantic consistency.
To address these challenges, {\method} integrates Semantic-Structural Distillation and Stage-Aware Allocation (SSD-SAA) for domain and schema grounding with a Critic--Reviser feedback loop for iterative diagnosis, targeted repair, and intent-level semantic verification. SSD-SAA refines semantic and structural knowledge and allocates it to the corresponding synthesis stages, while the feedback loop uses execution outcomes and diagnostic state transitions to progressively repair candidate programs. Together, these mechanisms turn analytical program generation from one-shot translation into a closed-loop process in which grounding, execution, diagnosis, repair, and verification are explicitly coordinated.
Our quantitative evaluation on Spider 2.0-Lite and BIRD-SQL Mini-Dev shows that {\method} achieves strong execution accuracy across multiple backbone models, SQL dialects, and query complexities. Under directly comparable settings, {\method} also improves token efficiency over existing iterative approaches. Additional analyses demonstrate the contributions of grounding, iterative refinement, and semantic verification, while the quantitative-finance case study illustrates the potential transfer of the framework beyond SQL-based analytics.

\section*{Acknowledgment}
This work was supported by the ``Pioneer'' R\&D Program of Zhejiang (2026C01019), and it was also supported by the National Natural Science Foundation of China (NSFC) under Grant No. 62572422.
GPT-5.6 Sol was used to assist with language polishing, grammar enhancement, and stylistic refinement of the manuscript.

\bibliographystyle{IEEEtran} 
\bibliography{refs}

\clearpage
\section{Appendix}
\subsection{Challenge Details}
As described in the Introduction, LLM analytical program synthesis mainly faces two challenges: \textbf{Domain and Schema Grounding}, \textbf{Execution Reliability and Semantic Consistency}.

\begin{figure}[b]
    \centering
    \includegraphics[width=\linewidth]{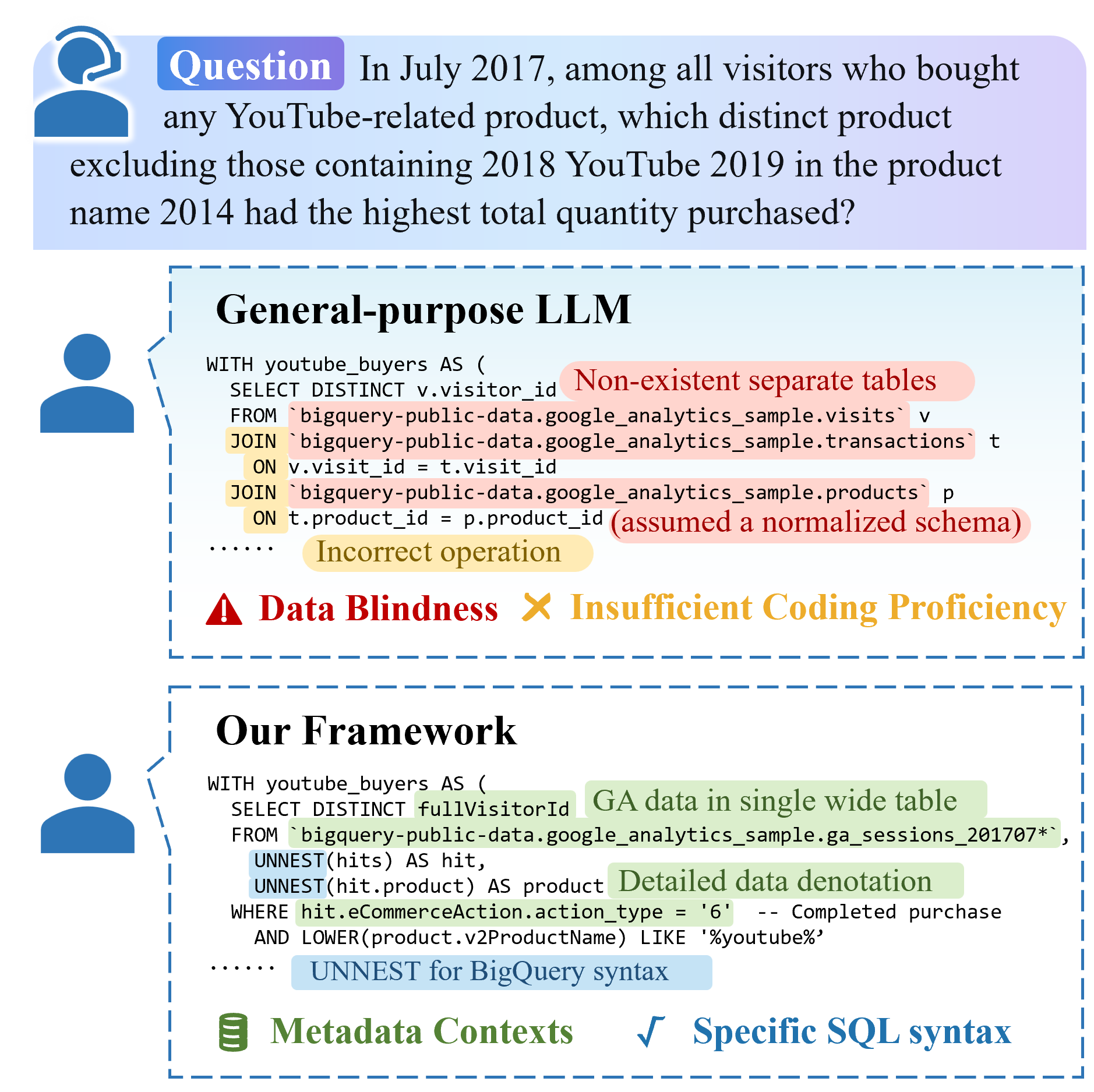}
    \caption{An Example of Challenges}
    \label{fig:challenges}
\end{figure}

The figure~\ref{fig:challenges} illustrates these two challenges.
\textbf{General-purpose LLM} produces an incorrect SQL snippet. It attempts to join \texttt{visits}, \texttt{transactions}, and \texttt{products} tables using inconsistent columns (\texttt{visitor\_id}, \texttt{visit\_id}, \texttt{product\_id}) and lacks proper handling of nested structures (e.g., \texttt{UNNEST} on hits and products). This directly exemplifies \textbf{C1}: the model fails to understand the actual BigQuery schema of the \texttt{google\_analytics\_sample} dataset, where purchase events are stored within repeated \texttt{hits} and \texttt{product} fields. As a result, the query is not only semantically wrong but also non-executable.
\textbf{Our Framework} demonstrates the correct approach: using \texttt{ga\_sessions\_201707*}, unnesting the fields of \texttt{hits} and \texttt{product}, filtering by \texttt{eCommerceAction.action\_type = '6'} (completed purchase), and applying a \texttt{LOWER(product.v2ProductName) LIKE '\%youtube\%'} condition. This respects the schema and maintains semantic integrity.

Thus, the figure highlights both challenges: (1) lack of schema awareness leads to syntactically flawed joins and missing unnest operations (C1); (2) even if a query were executable, preserving correct multi-step logic—such as distinguishing purchasers from all visitors, excluding specific product name patterns, and aggregating quantities—requires careful semantic consistency (C2), which single‑pass LLMs often violate.

\subsection{SSD-SAA Details}
Retrieval-augmented generation has emerged as a pivotal technique for mitigating common limitations of LLMs, such as hallucinations and outdated internal knowledge~\cite{gao2023retrieval, fan2024survey}. By dynamically integrating external knowledge sources, this mechanism enhances the factual accuracy of generated content and enables continuous knowledge updates without retraining the underlying model. 

\textit{Conceptual Semantics: }
Our design integrates \textbf{multiple knowledge sources}, employs dense search for better semantic matching, and incorporates pre-processing (e.g., chunking and indexing) as well as refinement steps such as similarity ranking and filtering.

The Spider 2.0 benchmark provides external knowledge \textbf{documents} for part of its questions~\cite{lei2024spider}. This is presented in \textbf{markdown documents} that define relevant domain terms, thereby offering the tested agent explanations for the questions.
However, most questions are not paired with such external knowledge. In the \textbf{Spider 2.0-lite} dataset, which contains 547 examples, only 107 are accompanied by documentation, resulting in a coverage rate of 19.56\%.

To compensate for the frequent lack of external knowledge, we introduce a Wikipedia searcher to supplement content. This can aid the LLM in better understanding tasks and deeper reasoning~\cite{chen2021evaluating, brown2020language}.

We first prompt the LLM to read the question and identify specialized terms it does not understand. The LLM is instructed to output these terms as a comma-separated list (e.g., \textit{"vaccine site, American Community Survey"} from the question: \textit{“Given the latest population estimates from the 2018 five-year American Community Survey, what is the number of vaccine sites per 1000 people for counties in California?”} ). Subsequently, the \textbf{WikipediaLoader} tool from LangChain~\cite{topsakal2023creating} is utilized to retrieve and download the entries for reference.
\textbf{Metadata Parsing:} While Spider 2.0-lite provides local Metadata, the complete dataset is primarily hosted on cloud database systems such as BigQuery and Snowflake~\cite{cao2024spider2}.

To handle the large volume of external documents (e.g., Wikipedia entries), we pre-process the text through a split-and-embed pipeline. Documents are first segmented into coherent chunks to minimize irrelevant content interference, then embedded and indexed using the \textbf{FAISS} (Facebook AI Similarity Search) vectorstore~\cite{douze2025faiss}. FAISS enables scalable and efficient approximate nearest neighbor search, which is essential for querying large-scale knowledge bases. During inference, the user’s natural language question is encoded into a dense vector, and a similarity search retrieves the top-k most relevant text passages.

\textit{Structural Schema: }
The Metadata for each database consists of two main components: Data Definition Language (DDL) statements in CSV files and a series of JSON summaries for every table. Each JSON summary details the table name, its Fully Qualified Name (FQN), column names, column data types, corresponding descriptions, and sample rows.

A notable challenge arises because many databases contain numerous homogeneous tables, which share an identical structure. For instance, the database \textsf{bigquery-public-data.google\_analytics\_sample} comprises 366 tables (e.g., \textsf{ga\_sessions\_20160923} through \textsf{ga\_sessions\_20170127}). To reduce informational redundancy and shorten the prompt length, we designed a specialized Metadata parsing mechanism. For homogeneous tables (e.g., those differentiated only by dates), the metadata is integrated, and only one representative table is described in detail. For heterogeneous tables, information for each table is presented and organized in a unified format.

A snippet of metadata parsing summary of the Snowflake database \texttt{\_1000\_GENOMES} is displayed as follows:

\lstset{
    basicstyle=\ttfamily\footnotesize, 
    breaklines=true,                  
    frame=none,                       
    backgroundcolor=\color{gray!10},  
    keywordstyle=\color{blue},        
    commentstyle=\color{green!50!black}, 
    stringstyle=\color{red},          
    numbers=left,                     
    numberstyle=\tiny\color{gray},    
    xleftmargin=2em,                  
    framexleftmargin=1.5em,            
    upquote=true,              
    columns=fullflexible 
}

\tcbset{
    enhanced,
    breakable,
    colback=gray!10, 
    colframe=gray!50, 
    boxrule=0.5pt, 
    arc=2mm, 
    left=1mm, 
    right=1mm, 
    top=1mm, 
    bottom=1mm 
}

\noindent \textbf{The schema has 3 tables in total.}
\begin{tcolorbox}[title=SAMPLE\_INFO DDL]
\begin{lstlisting}[language=SQL]
create or replace TABLE SAMPLE_INFO (
"Sample" VARCHAR(16777216),
"Family_ID" VARCHAR(16777216),
"Population" VARCHAR(16777216),
"Population_Description" VARCHAR(16777216),
"Gender" VARCHAR(16777216),
"Relationship" VARCHAR(16777216),
"Unexpected_Parent_Child" VARCHAR(16777216),
"Non_Paternity" VARCHAR(16777216),
"Siblings" VARCHAR(16777216),
"Grandparents" VARCHAR(16777216),
"Avuncular" VARCHAR(16777216),
"Half_Siblings" VARCHAR(16777216),
"Unknown_Second_Order" VARCHAR(16777216),
"Third_Order" VARCHAR(16777216),
"In_Low_Coverage_Pilot" BOOLEAN,
"LC_Pilot_Platforms" VARCHAR(16777216),
"LC_Pilot_Centers" VARCHAR(16777216),
"In_High_Coverage_Pilot" BOOLEAN,
"HC_Pilot_Platforms" VARCHAR(16777216),
"HC_Pilot_Centers" VARCHAR(16777216),
"In_Exon_Targetted_Pilot" BOOLEAN,
"ET_Pilot_Platforms" VARCHAR(16777216),
"ET_Pilot_Centers" VARCHAR(16777216),
"Has_Sequence_in_Phase1" BOOLEAN,
"Phase1_LC_Platform" VARCHAR(16777216),
"Phase1_LC_Centers" VARCHAR(16777216),
"Phase1_E_Platform" VARCHAR(16777216),
"Phase1_E_Centers" VARCHAR(16777216),
"In_Phase1_Integrated_Variant_Set" BOOLEAN,
"Has_Phase1_chrY_SNPS" BOOLEAN,
"Has_phase1_chrY_Deletions" BOOLEAN,
"Has_phase1_chrMT_SNPs" BOOLEAN,
"Main_project_LC_Centers" VARCHAR(16777216),
"Main_project_LC_platform" VARCHAR(16777216),
"Total_LC_Sequence" FLOAT,
"LC_Non_Duplicated_Aligned_Coverage" FLOAT,
"Main_Project_E_Centers" VARCHAR(16777216),
"Main_Project_E_Platform" VARCHAR(16777216),
"Total_Exome_Sequence" FLOAT,
"X_Targets_Covered_to_20x_or_greater" FLOAT,
"VerifyBam_E_Omni_Free" FLOAT,
"VerifyBam_E_Affy_Free" FLOAT,
"VerifyBam_E_Omni_Chip" FLOAT,
"VerifyBam_E_Affy_Chip" FLOAT,
"VerifyBam_LC_Omni_Free" FLOAT,
"VerifyBam_LC_Affy_Free" FLOAT,
"VerifyBam_LC_Omni_Chip" FLOAT,
"VerifyBam_LC_Affy_Chip" FLOAT,
"LC_Indel_Ratio" FLOAT,
"E_Indel_Ratio" FLOAT,
"LC_Passed_QC" BOOLEAN,
"E_Passed_QC" BOOLEAN,
"In_Final_Phase_Variant_Calling" BOOLEAN,
"Has_Omni_Genotypes" BOOLEAN,
"Has_Axiom_Genotypes" BOOLEAN,
"Has_Affy_6_0_Genotypes" BOOLEAN,
"Has_Exome_LOF_Genotypes" BOOLEAN,
"EBV_Coverage" FLOAT,
"DNA_Source_from_Coriell" VARCHAR(16777216),
"Has_Sequence_from_Blood_in_Index" BOOLEAN,
"Super_Population" VARCHAR(16777216),
"Super_Population_Description" VARCHAR(16777216)
);
\end{lstlisting}
\end{tcolorbox}

\noindent \textbf{Sample Row:}
\begin{tcolorbox}
\begin{lstlisting}[language=json]
{
"Sample": "HG02974",
"Family_ID": "NG23",
"Population": "ESN",
"Population_Description": "Esan in Nigeria",
"Gender": "female",
"Relationship": "mother",
"Unexpected_Parent_Child": "",
"Non_Paternity": "",
"Siblings": "",
"Grandparents": "",
"Avuncular": "",
"Half_Siblings": "",
"Unknown_Second_Order": "",
"Third_Order": "",
"In_Low_Coverage_Pilot": null,
"LC_Pilot_Platforms": "",
"LC_Pilot_Centers": "",
"In_High_Coverage_Pilot": null,
"HC_Pilot_Platforms": "",
"HC_Pilot_Centers": "",
"In_Exon_Targetted_Pilot": null,
"ET_Pilot_Platforms": "",
"ET_Pilot_Centers": "",
"Has_Sequence_in_Phase1": null,
"Phase1_LC_Platform": "",
"Phase1_LC_Centers": "",
"Phase1_E_Platform": "",
"Phase1_E_Centers": "",
"In_Phase1_Integrated_Variant_Set": null,
"Has_Phase1_chrY_SNPS": null,
"Has_phase1_chrY_Deletions": null,
"Has_phase1_chrMT_SNPs": null,
"Main_project_LC_Centers": "BI",
"Main_project_LC_platform": "ILLUMINA",
"Total_LC_Sequence": 32595580924.0,
"LC_Non_Duplicated_Aligned_Coverage": 10.17,
"Main_Project_E_Centers": "BI",
"Main_Project_E_Platform": "ILLUMINA",
"Total_Exome_Sequence": 7576615180.0,
"X_Targets_Covered_to_20x_or_greater": 0.917355,
"VerifyBam_E_Omni_Free": 0.00487,
"VerifyBam_E_Affy_Free": 0.00541,
"VerifyBam_E_Omni_Chip": null,
"VerifyBam_E_Affy_Chip": 0.00543,
"VerifyBam_LC_Omni_Free": 0.00263,
"VerifyBam_LC_Affy_Free": 0.0038,
"VerifyBam_LC_Omni_Chip": null,
"VerifyBam_LC_Affy_Chip": 0.00391,
"LC_Indel_Ratio": 1.3,
"E_Indel_Ratio": 1.09,
"LC_Passed_QC": true,
"E_Passed_QC": true,
"In_Final_Phase_Variant_Calling": true,
"Has_Omni_Genotypes": null,
"Has_Axiom_Genotypes": null,
"Has_Affy_6_0_Genotypes": true,
"Has_Exome_LOF_Genotypes": null,
"EBV_Coverage": 92.29,
"DNA_Source_from_Coriell": "LCL",
"Has_Sequence_from_Blood_in_Index": null,
"Super_Population": "AFR",
"Super_Population_Description": "African"
}
\end{lstlisting}
\end{tcolorbox}

\begin{tcolorbox}[title=PEDIGREE DDL]
\begin{lstlisting}[language=SQL]
create or replace TABLE PEDIGREE (
"Family_ID" VARCHAR(16777216),
"Individual_ID" VARCHAR(16777216),
"Paternal_ID" VARCHAR(16777216),
"Maternal_ID" VARCHAR(16777216),
"Gender" NUMBER(38,0),
"Phenotype" NUMBER(38,0),
"Population" VARCHAR(16777216),
"Relationship" VARCHAR(16777216),
"Siblings" VARCHAR(16777216),
"Second_Order" VARCHAR(16777216),
"Third_Order" VARCHAR(16777216),
"Other_Comments" VARCHAR(16777216)
);
\end{lstlisting}
\end{tcolorbox}

\begin{tcolorbox}[title=VARIANTS DDL]
\begin{lstlisting}[language=SQL]
create or replace TABLE VARIANTS (
"reference_name" VARCHAR(16777216),
"start" NUMBER(38,0),
"end" NUMBER(38,0),
"reference_bases" VARCHAR(16777216),
"alternate_bases" VARIANT,
"quality" FLOAT,
"filter" VARIANT,
"names" VARIANT,
"call" VARIANT,
AA VARCHAR(16777216),
AC VARIANT,
AC1 NUMBER(38,0),
AF FLOAT,
AF1 FLOAT,
AFR_AF FLOAT,
AMR_AF FLOAT,
AN NUMBER(38,0),
ASN_AF FLOAT,
AVGPOST FLOAT,
CIEND VARIANT,
CIPOS VARIANT,
DP NUMBER(38,0),
DP4 VARIANT,
ERATE FLOAT,
EUR_AF FLOAT,
FQ FLOAT,
G3 VARIANT,
HOMLEN NUMBER(38,0),
HOMSEQ VARCHAR(16777216),
HWE FLOAT,
LDAF FLOAT,
MQ NUMBER(38,0),
PV4 VARIANT,
RSQ FLOAT,
SNPSOURCE VARIANT,
SOURCE VARIANT,
SVLEN NUMBER(38,0),
SVTYPE VARCHAR(16777216),
THETA FLOAT,
VT VARCHAR(16777216)
);
\end{lstlisting}
\end{tcolorbox}

These passages, together with the parsed Metadata, are assembled into an augmented prompt, as \textbf{Input-Layer Integration}~\cite{fan2024survey}. Retrieved Wikipedia passages and documentation are combined with the parsed metadata and incorporated into the agent's prompt. This process synthesizes a comprehensive background for the task, enabling the agent to ground its interpretation in relevant external knowledge and significantly reduce schema and terminological hallucinations, guiding it to generate more accurate and schema-aware SQL queries~\cite{brown2020language}.

\begin{figure}[htbp]
    \centering
    \includegraphics[width=\linewidth]{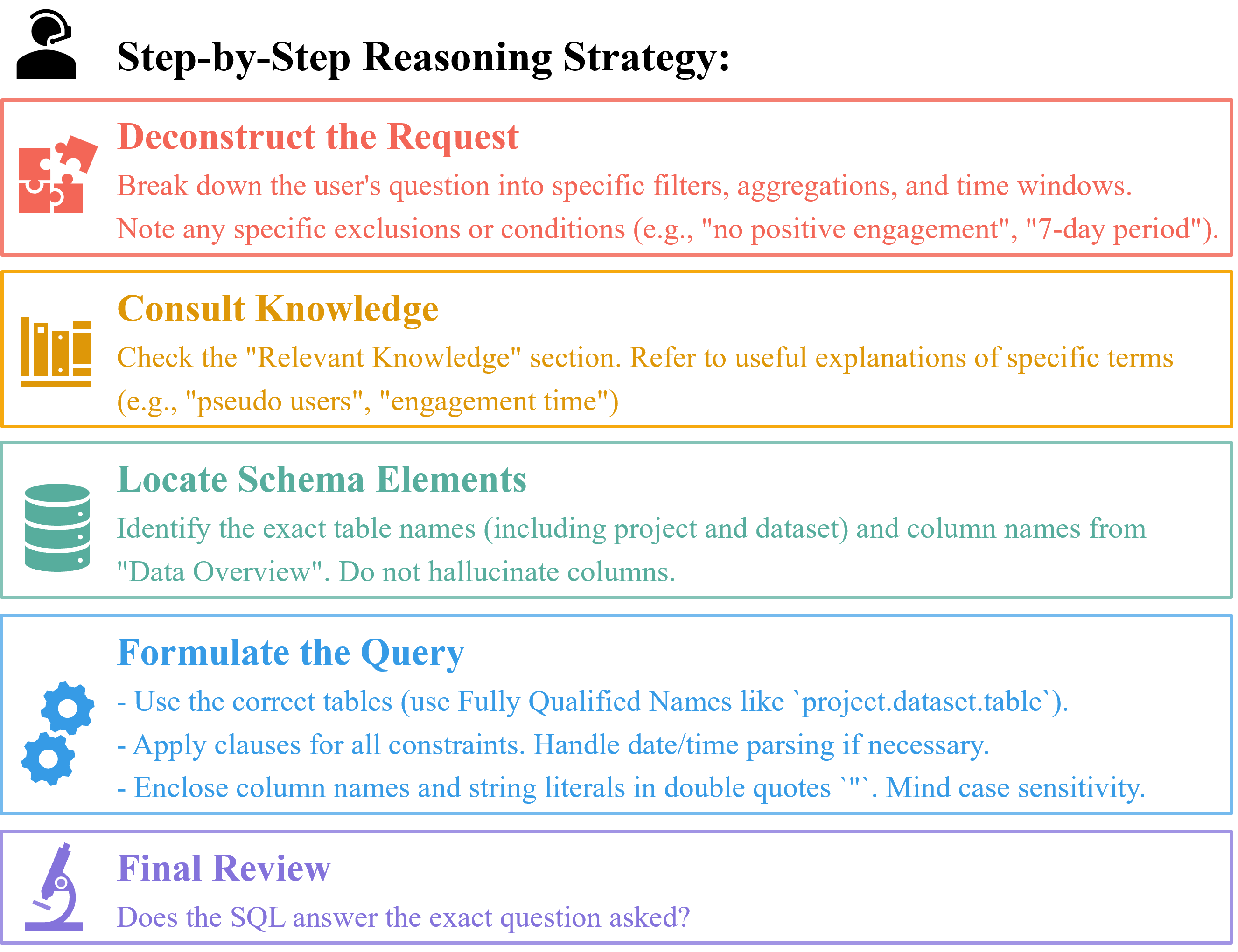}
    \caption{Stage-aware Reasoning}
    \label{fig:stage}
\end{figure}

\textit{Stage-aware Allocation: }
To enhance stage-aware reasoning, the agent follows a carefully designed prompting strategy. By providing the modularized reasoning decomposition, LLMs can be guided to solve tasks by eliciting intermediate logical stages, thereby offering both better accuracy and a degree of explainability~\cite{wei2022chain, kojima2022large}. In our implementation, the model is guided through a systematic reasoning workflow, illustrated in Figure~\ref{fig:stage}. It incorporates the key parts below: 

1. \textbf{Deconstruct:} Break the user request into core components such as filters, aggregations, and desired outputs. 

2. \textbf{Consult:} Leverage retrieved domain knowledge to clarify ambiguous terms and contextualize the question.  

3. \textbf{Locate:} Map natural language phrases precisely to database schema elements (tables, columns) using metadata, avoiding hallucinated references.

4. \textbf{Formulate:} Compose the SQL query with strict syntax adherence, including fully qualified names, constraints, type‑aware formatting, and proper string escaping.

5. \textbf{Review:} Self‑verify the logical alignment between the query and the original question.

This step‑wise approach simulates the staged problem‑solving of a human expert, allowing the agent to decompose complex tasks, fully utilize the provided context, and produce more reliable, executable code.

\subsection{Agent Prompt Design}

In this section, we detail the specific prompt designs for the three core agents in our framework. We present representative prompts with clear hierarchical structures and modular components to ensure reproducibility and transparency.

 \subsubsection{Generator Agent}

The Generator Agent is responsible for creating initial code based on user requests, database schemas, and contextual knowledge. It employs comprehensive reasoning strategies to ensure accurate and executable code generation while maintaining strict schema alignment.

 \paragraph{Core Mechanism}

The agent synthesizes code through a structured five-step reasoning process that decomposes user requests, consults knowledge bases, locates schema elements, formulates queries, and performs final validation. The process emphasizes the use of fully qualified table names and proper column referencing to prevent common errors in BigQuery environments.

\begin{promptbox}{Prompt for Generator Agent }
\textbf{Role:} You are an excellent SQL expert.
\\
\textbf{Task:} Write a SQL query to answer the user's request based on the provided database schema.
\\
\\
\textbf{Contextual Inputs:}
\begin{itemize}
    \item User request: \{query\}
    \item Data Overview: \{data\_context\}
    \item Relevant knowledge: \{context\}
\end{itemize}

\textbf{Staged Reasoning Strategy:}
\begin{enumerate}
    \item \textbf{Deconstruct the Request:} Break down the user's question into specific filters, aggregations, and time windows. Note any specific exclusions or conditions (e.g., "no positive engagement", "7-day period").
    \item \textbf{Consult Knowledge:} Check the "Relevant Knowledge" section. Refer to useful explanations of specific terms (e.g., "pseudo users", "engagement time").
    \item \textbf{Locate Schema Elements:} Identify the exact table names (including project and dataset) and column names from "Data Overview". Do not hallucinate columns.
    \item \textbf{Formulate the Query:}
    \begin{itemize}
        \item Use the correct tables.
        \item Apply clauses for all constraints.
        \item Handle date/time parsing if necessary.
    \end{itemize}
    \item \textbf{Final Review:} Does the SQL answer the exact question asked?
\end{enumerate}

\textbf{Output Requirements:}
\begin{enumerate}
    \item First, provide a short Analysis of the question.
    \item Then, output the SQL inside a Markdown code block.
\end{enumerate}

\textbf{Output Format:}
\\
Analysis: [Your brief thought process]
\\
\\
\texttt{```sql}
\\
SELECT (your SQL clause)
\\
FROM \{fqn\}
\\
(your SQL clause)
\\
\texttt{```}
\\
(Replace with the actual table and column names using Fully Qualified Name)
\end{promptbox}

\paragraph{Error Handling and Safety Features}

The implementation includes robust error-handling mechanisms:
\begin{itemize}
    \item \textbf{Content Filtering:} Automatic retry logic when API content filters are triggered, omitting external knowledge contexts while preserving core functionality
    \item \textbf{Context Truncation:} Intelligent truncation of lengthy data contexts (>350,000 characters) with clear indication of truncation to maintain prompt size limits
    \item \textbf{FQN Resolution:} Automated extraction of fully qualified table names from available schema information
\end{itemize}

\subsubsection{Critic Agent}

The Critic Agent adopts the \textbf{Delta-guided Memory Management} mechanism, serving as a meta-cognitive component, which monitors the debugging life-cycle, evaluates correction effectiveness, and provides strategic guidance for overcoming persistent challenges. It functions as a quality assurance layer that prevents debugging loops and promotes systematic problem-solving.

\paragraph{Reflection and Progress Assessment}

The agent implements a comparative analysis framework that evaluates whether previous fixes successfully resolved errors or created new issues. Tracking the evolution of errors and code changes, it provides actionable insights for navigating complex debugging scenarios.

\begin{promptbox}{Prompt for Critic Agent}
\textbf{Role:} You are an expert Reviewer.
\\
\textbf{Goal:} Check the debugging progress and provide actionable suggestions for the next step.
\\
\\
\textbf{Reflection Context:}
\begin{itemize}
    \item Previous Error History: \{history\}
    \item Last Encountered Error: \{last\_error\}
    \item Last SQL Code: \{last\_code\}
    \item Current Error: \{current\_error\}
    \item Current SQL Code: \{current\_code\}
\end{itemize}

\textbf{Reflection Process (Chain of Thought):}
\begin{enumerate}
    \item \textbf{Compare Errors:} Did the "Last Encountered Error" disappear in the "Current Error"? 
    \begin{itemize}
        \item If YES: The previous fix was successful (or at least moved past that bug).
        \item If NO (Same Error): The previous fix FAILED. We are stuck in a loop.
    \end{itemize}
    \item \textbf{Formulate Advice:} 
    \begin{itemize}
        \item If progressing: Summarize the error. Compare the last SQL code with the current SQL code, and describe the solution to it.
        \item If stuck: Suggest a different approach. You can also refer to the previous error history for some inspiration.
    \end{itemize}
\end{enumerate}

\textbf{Output Format:}
\\
Current Code [\{current\_code\}]. Current Error[\{current\_error\}].
\\
Previous fix [Worked/Failed]. Suggestion: [Action].
\end{promptbox}

 \paragraph{Strategic Decision Support}

The Critic Agent provides binary classification of fix effectiveness ("Worked"/"Failed") coupled with concrete action suggestions. This clear directive format enables the debugging system to escape local optima and explore alternative correction strategies when standard approaches prove ineffective.

\subsubsection{Reviser Agent}

The Reviser Agent specializes in diagnosing and fixing execution errors through systematic debugging strategies. It operates as a corrective layer that analyzes error messages, historical error patterns, and reference hints to produce corrected code.

\paragraph{Debugging Methodology}

The agent employs a four-phase debugging strategy that classifies errors, locates fault points, formulates targeted fixes, and ensures logical consistency. This structured approach prevents error recurrence and promotes learning from historical debugging patterns.

\begin{promptbox}{Prompt for Reviser Agent }
\textbf{Role:} You are an excellent SQL expert.
\\
\textbf{Task:} Correct the SQL based on the error message after the previous SQL query failed to execute.
\\
\\
\textbf{Debugging Context:}
\begin{itemize}
    \item User request: \{query\}
    \item Latest executed SQL: \{SQL\}
    \item Error: \{error\}
    \item Error history: \{error\_history\}
    \item Reference / Hints: \{reference\}
\end{itemize}

\textbf{Step-by-Step Debugging Strategy (Chain of Thought):}
\begin{enumerate}
    \item \textbf{Analyze the Error:} Read the error message carefully. Is it a syntax error, a schema mismatch, or a logic issue?
    \item \textbf{Locate the Fault:} Identify the exact line or clause in the SQL that caused the error.
    \begin{itemize}
        \item If it's a syntax error, check the command and function usage.
        \item If it's a schema mismatch, check the correct FQN for tables and correct case-sensitive columns.
        \item If it's a logic issue, review the user request constraints (e.g., date ranges, filters) carefully.
    \end{itemize}
    \item \textbf{Formulate a Fix:} Make use of your knowledge and the reference/hints to fix the issues.
    \item \textbf{Final Refinement:} Ensure the rest of the query (filters, logic) remains consistent with the User Request.
    \begin{itemize}
        \item For syntax/schema: applying the correct function or name.
        \item For logic: consider rewriting the CTEs or subqueries to align with the correct logic.
    \end{itemize}
\end{enumerate}

\textbf{Output Requirements:}
\begin{enumerate}
    \item First, provide a short Analysis of why the error happened and how you will fix it.
    \item Then, output the Corrected SQL inside a Markdown code block.
\end{enumerate}

\textbf{Output Format:}
\\
Analysis: [Your brief debugging thought process]
\\
\\
\texttt{```sql}
\\
SELECT (your SQL clause)
\\
FROM \{fqn\}
\\
(your SQL clause)
\\
\texttt{```}
\\
(Replace with the actual table and column names using FQN)
\end{promptbox}

\paragraph{Historical Error Learning}

The Reviser Agent maintains and utilizes an error history context offered by the Critic Agent to identify recurring error patterns and prevent previously encountered issues. This historical awareness enables more intelligent debugging that evolves based on accumulated correction experiences.

\subsubsection{System Integration and Workflow}

The three agents operate in a coordinated pipeline:

\begin{enumerate}
    \item \textbf{Generator Agent} produces initial SQL queries with comprehensive reasoning
    \item \textbf{Reviser Agent} intervenes when execution errors occur, applying systematic debugging
    \item \textbf{Critic Agent} monitors the debugging trajectory, providing meta-level guidance
\end{enumerate}

This hierarchical structure ensures that SQL generation benefits from both immediate error correction and long-term debugging strategy optimization. The prompt designs emphasize consistency in output formatting, use of fully qualified names, and adherence to BigQuery-specific syntax conventions, creating a robust framework for automated SQL query generation and refinement.

\subsection{Evaluation Metric Details}

Let $Y_n$ and $\hat{Y}_n$ denote the ground-truth and
predicted SQL queries for the $n$-th example, respectively,
and let $V_n$ and $\hat{V}_n$ denote their corresponding
execution results. Let $N$ denote the number of examples
in the evaluation set.

\paragraph{Execution Accuracy (EX)}
EX measures the proportion of examples for which the
predicted SQL query returns results equivalent to those
of the ground-truth query~\cite{qin2022survey}.
For BIRD and Spider 2.0-Lite, it is computed
as~\cite{li2023can, lei2024spider}:
\begin{equation*}
    \mathrm{EX}
    = \frac{1}{N}
      \sum_{n=1}^{N}
      \mathbbm{1}(V_n, \hat{V}_n),
\end{equation*}
where $\mathbbm{1}(\cdot,\cdot)$ is an indicator function:
\begin{equation*}
    \mathbbm{1}(V, \hat{V})
    =
    \begin{cases}
        1, & V = \hat{V}, \\
        0, & \text{otherwise}.
    \end{cases}
\end{equation*}
Here, equality is determined by the result-comparison
rules of the corresponding benchmark. Predictions
that fail to execute receive a score of zero.

\paragraph{Valid Efficiency Score (VES)}
VES evaluates execution efficiency while accounting
for execution correctness. Only predictions whose
execution results match the ground-truth results
contribute to the score~\cite{li2023can}:
\begin{equation*}
    \begin{aligned}
        \mathrm{VES}
        &= \frac{1}{N}
           \sum_{n=1}^{N}
           \mathbbm{1}(V_n, \hat{V}_n)\,\rho_n, \\
        \rho_n
        &= \sqrt{\frac{T(Y_n)}{T(\hat{Y}_n)}}.
    \end{aligned}
\end{equation*}
Here, $T(\cdot)$ denotes the execution time of a SQL
query measured under the same evaluation environment,
and $\rho_n$ represents the relative execution
efficiency of the predicted query.
Incorrect or non-executable predictions contribute
zero to VES.

\paragraph{Soft-F1}
Soft-F1 measures partial agreement between the predicted
and ground-truth execution results at the cell level.
Let $\mathcal{C}(V_n)$ and $\mathcal{C}(\hat{V}_n)$
denote the sets of cell values in the ground-truth and
predicted results, respectively.
When both result sets are nonempty, the instance-level
precision $P_n$ and recall $R_n$ are defined as:
\begin{equation*}
    \begin{aligned}
        P_n
        &= \frac{
            |\mathcal{C}(V_n)
            \cap \mathcal{C}(\hat{V}_n)|
        }{
            |\mathcal{C}(\hat{V}_n)|
        }, \\[4pt]
        R_n
        &= \frac{
            |\mathcal{C}(V_n)
            \cap \mathcal{C}(\hat{V}_n)|
        }{
            |\mathcal{C}(V_n)|
        }.
    \end{aligned}
\end{equation*}
Here, $|\cdot|$ denotes set cardinality.
When both result sets are empty, we set $P_n=R_n=1$;
when exactly one is empty, we set $P_n=R_n=0$.
The instance-level Soft-F1 score is:
\begin{equation*}
    F_n =
    \begin{cases}
        \displaystyle
        \frac{2P_nR_n}{P_n+R_n},
        & P_n+R_n>0, \\[6pt]
        0, & \text{otherwise}.
    \end{cases}
\end{equation*}
The overall score is the average across all
evaluation examples:
\begin{equation*}
    \mathrm{Soft\text{-}F1}
    = \frac{1}{N}\sum_{n=1}^{N} F_n.
\end{equation*}

\subsection{Quantitative Application Case Supplements}
\label{sec:appendix-case-quantitative}

\subsubsection{Similar Challenges}
In the field of quantitative investment and financial computing, LLMs are confronted with similar challenges mentioned in Section ~\ref{section::introduction}:

\textbf{Financial Domain Semantics:} General-purpose LLMs typically lack deep semantic understanding of \textbf{specialized operators} and functions in alpha factor construction (e.g. \texttt{delta}, \texttt{delay}, \texttt{adv\{d\}}, \texttt{industry neutralization})~\cite{pahune2023several}. Also, financial data are usually stored as large-scale tabular structures (\texttt{CSV}, \texttt{DataFrames}) with thousands of rows (e.g., daily trading records), which is hard for existing LLMs to handle~\cite{liu2024rethinking, gardner2024large}. (Challenge 1)

\textbf{Unreliable Numerical Operations:}  Factor calculations usually involve \textit{rolling windows}, \textit{cross-sectional rankings}, and \textit{multi-level nesting}, where LLMs tend to lose intermediate states~\cite{liu2024rethinking} and make computational errors. A snippet of \texttt{000006.SZ} trading data is shown in Table~\ref{tab:stock_data}.(Challenge 2)

\begin{table}[htbp]
  \centering
  \caption{Sample of Stock Trading Data (000006.SZ)}
  \label{tab:stock_data}
  \begin{tabular}{>{\columncolor{gray!15}}lp{0.5cm}p{0.5cm}p{0.5cm}p{0.5cm}p{0.5cm}r}
    \toprule
    \textbf{Date} & \textbf{Open} & \textbf{High} & \textbf{Low} & \textbf{Close} & $\dots$ & \textbf{Industry} \\
    \midrule
    2009-01-05 & 55.09 & 56.84 & 54.76 & 56.73 & $\dots$ & Real Estate \\
    2009-01-06 & 56.73 & 59.03 & 56.18 & 59.03 & $\dots$ & Real Estate \\
    \multicolumn{1}{c}{$\vdots$} & $\vdots$ & $\vdots$ & $\vdots$ & $\vdots$ & $\ddots$ & $\vdots$ \\
    2025-08-15 & 285.33 & 294.87 & 283.35 & 291.29 & $\dots$ & Real Estate \\
    \bottomrule
  \end{tabular}
\end{table}

The Quantitative Analysis Agent system is designed to generate, execute, and debug Python scripts for quantitative financial analysis tasks. The system consists of three interconnected agents: \textbf{Generator Agent} for initial code generation, \textbf{Reviser Agent} for error correction, and \textbf{Critic Agent} for debugging progress monitoring.

\subsubsection{Domain Knowledge Injection}

The system incorporates extensive domain knowledge for quantitative finance calculations, providing definitions for financial metrics and mathematical operators.

\begin{promptbox}{Quantitative Finance Domain Knowledge Context}
\textbf{Input data definitions:}
\begin{itemize}
    \item \texttt{returns} = daily close-to-close returns
    \item \texttt{open, close, high, low, volume} = standard definitions for daily price and volume data
    \item \texttt{vwap} = daily volume-weighted average price
    \item \texttt{cap} = market cap
    \item \texttt{adv\{d\}} = average daily dollar volume for the past d days
    \item \texttt{IndClass} = a generic placeholder for a binary industry classification such as GICS, BICS, NAICS, SIC, etc.
\end{itemize}

\textbf{Functions and operators:}
(Below, "``\{ \}'' stands for a placeholder)

\begin{itemize}
    \item \texttt{abs(x), log(x), sign(x)} = standard definitions; same for the operators "+", "-", "*", "/", ">", "<", "==", "||", "x ? y : z"
    \item \texttt{rank(x)} = cross-sectional rank
    \item \texttt{delay(x, d)} = value of x d days ago
    \item \texttt{correlation(x, y, d)} = time-series correlation of x and y for the past d days
    \item \texttt{covariance(x, y, d)} = time-series covariance of x and y for the past d days
    \item \texttt{scale(x, a)} = rescaled x such that sum(abs(x)) = a (the default is a = 1)
    \item \texttt{delta(x, d)} = today's value of x minus the value of x d days ago
    \item \texttt{signedpower(x, a)} = x\^{}a
    \item \texttt{decay\_linear(x, d)} = weighted moving average over the past d days with linearly decaying weights d, d - 1, …, 1 (rescaled to sum up to 1)
    \item \texttt{indneutralize(x, g)} = x cross-sectionally neutralized against groups g (subindustries, industries, sectors, etc.)
    \item \texttt{ts\_\{O\}(x, d)} = operator O applied across the time series for the past d days
    \item \texttt{ts\_min(x, d)} = time-series min over the past d days
    \item \texttt{ts\_max(x, d)} = time-series max over the past d days
    \item \texttt{ts\_argmax(x, d)} = which day ts\_max(x, d) occurred on
    \item \texttt{ts\_argmin(x, d)} = which day ts\_min(x, d) occurred on
    \item \texttt{ts\_rank(x, d)} = time-series rank in the past d days
    \item \texttt{min(x, d)} = ts\_min(x, d)
    \item \texttt{max(x, d)} = ts\_max(x, d)
    \item \texttt{sum(x, d)} = time-series sum over the past d days
    \item \texttt{product(x, d)} = time-series product over the past d days
    \item \texttt{stddev(x, d)} = moving time-series standard deviation over the past d days
\end{itemize}
\end{promptbox}

\subsubsection{Synthesis Pipeline}
The whole program generation process is shown in the Figure~\ref{fig:pipeline}. The details are described as follows.

\paragraph{Initial Code Generation (Generator Agent)}

The Generator Agent is responsible for creating initial Python scripts for quantitative analysis tasks based on user requests, data schemas, and domain knowledge. It employs structured implementation strategies to ensure accurate and efficient code generation.

\textit{Initialization Prompt: } 
The core mechanism involves guiding the LLM through a four-step implementation strategy focusing on vectorized operations and proper data handling.

\begin{promptbox}{Prompt for Generator Agent (Abbreviated)}
\textbf{Role:} You are an expert Quantitative Developer and Data Scientist.

\textbf{Task:} Write a Python script to perform quantitative analysis and calculation based on the user's request and provided CSV data.

\textbf{Context:}
\begin{itemize}
    \item User request: \{query\}
    \item Data Overview (File paths \& Columns): \{data\_context\}
    \item Relevant Domain Knowledge (Formulas/Definitions): \{context\}
\end{itemize}
......
\end{promptbox}

\paragraph{Code Error Correction (Reviser Agent)}

The Reviser Agent specializes in diagnosing and fixing Python execution errors through systematic debugging strategies. It operates in two modes: error correction mode and verification mode.

\textit{Error Correction Prompt:} When the code fails to execute, the agent analyzes error messages and applies targeted fixes.

\begin{promptbox}{Prompt for Reviser Agent - Error Correction (Abbreviated)}
\textbf{Role:} You are an expert Python Quantitative Developer.

\textbf{Task:} Fix the code based on the error message from previous execution.
\\
......
\end{promptbox}

\textit{Verification Prompt:} When the code executes successfully, the agent verifies if the logic and results match the user's request.

\begin{promptbox}{Prompt for Reviser Agent - Verification (Abbreviated)}
\textbf{Role:} You are an expert Quantitative Reviewer.

\textbf{Task:} Verify if the logic and results match the User's Request.
......
\end{promptbox}

\paragraph{Delta-guided Memory (Critic Agent)}

The Critic Agent serves as a meta-cognitive component that monitors the debugging life-cycle, evaluates correction effectiveness, and provides strategic guidance.

\textit{Reflection Prompt:} The agent analyzes debugging progress and suggests next steps based on error patterns.

\begin{promptbox}{Prompt for Critic Agent (Abbreviated)}
\textbf{Role:} You are an expert Reviewer.

\textbf{Goal:} Check debugging progress and provide actionable suggestions.
......
\end{promptbox}

\subsubsection{Program Pipeline for Factor computing}
The generated code for factor\#36 (in WorldQuant’s 101 Alphas~\cite{kakushadze2016101}) calculation is implemented as follows.

\noindent\textbf{Code Structure Summary:}
\begin{enumerate}
    \item \textcolor{blue}{\textbf{Data Loading}} (\texttt{lines 1-19}): Load multiple CSV files and merge into a single DataFrame
    \item \textcolor{blue}{\textbf{Preprocessing}} (\texttt{lines 22-33}): Calculate daily returns, VWAP, and 20-day average dollar volume
    \item \textcolor{blue}{\textbf{Time Series Ranking Function}} (\texttt{lines 36-40}): Define the Ts\_Rank helper function
    \item \textcolor{blue}{\textbf{Alpha Factor Calculation}} (\texttt{lines 43-82}): Compute the 5 Alpha factor components
    \item \textcolor{blue}{\textbf{Result Aggregation}} (\texttt{lines 85-94}): Combine components and convert to pivot table format
\end{enumerate}

\noindent\textbf{Key Calculations:}
\begin{itemize}
    \item \textcolor{green!60!black}{\texttt{ts\_rank}}: Rolling time series ranking function
    \item \textcolor{green!60!black}{\texttt{groupby().transform()}}: Group-wise rolling calculations
    \item \textcolor{green!60!black}{\texttt{pct=True}}: Percentage ranking computation
    \item \textcolor{green!60!black}{\texttt{rolling().corr()}}: Rolling correlation coefficient calculation
\end{itemize}

\begin{lstlisting}[style=pythonstyle]
import os
import pandas as pd
import numpy as np

# 1. Load all stock data
data_dir = "./stock_data"
all_files = [f for f in os.listdir(data_dir) if f.endswith('.csv')]
dfs = []
for file in all_files:
    code = file.replace('.csv', '')
    df = pd.read_csv(os.path.join(data_
    dir, file), dtype={'date': str})
    df['code'] = code
    dfs.append(df)
df = pd.concat(dfs, ignore_index=True)

# Convert date to datetime and set as index for easier manipulation
df['date'] = pd.to_datetime(df['date'])
df.sort_values(['code', 'date'], inplace=True)

# 2. Preprocessing: ensure numeric columns are numeric (they already are per schema)
# Compute returns: close-to-close daily returns
df['returns'] = df.groupby('code')['clo
se'].pct_change()

# Compute vwap: volume-weighted average price (using amount and volume)
# vwap = amount / volume (if volume > 0)
df['vwap'] = df['amount'] / df['volume'].replace(0, np.nan)

# Compute adv20: 20-day average dollar volume (amount)
df['adv20'] = df.groupby('code')['amount']
.transform( lambda x: x.rolling(20, min_periods=15).mean())

# 3. Calculate components of the alpha formula
# Helper function for time-series rank (ts_Rank)
def ts_rank(series, window):
    # Rolling time-series rank: rank over the window, scaled to [0,1]
    return series.rolling(window, min_periods=window//2).apply(
        lambda x: pd.Series(x).rank(pct=True)
        .iloc[-1], raw=False
    )

# Component 1: 2.21 * rank(correlation((close - open), delay(volume, 1), 15))
df['close_open'] = df['close'] - df['open']
df['volume_lag1'] = df.groupby('code')
['volume'].shift(1)
# Rolling correlation over 15 days
df['corr_co'] = df.groupby('code').apply(lambda x:
x['close_open'].rolling(15, min_periods
=10).corr(x['volume_lag1'])
).reset_index(level=0, drop=True)
# Cross-sectional rank of correlation
df['rank_corr'] = df.groupby('date')
['corr_co'].rank(pct=True)
comp1 = 2.21 * df['rank_corr']

# Component 2: 0.7 * rank((open - close))
df['open_close'] = df['open'] - df['close']
df['rank_oc'] = df.groupby('date')['open_close'].
rank(pct=True)
comp2 = 0.7 * df['rank_oc']

# Component 3: 0.73 * rank(ts_Rank(delay((-1 * returns), 6), 5))
df['neg_returns'] = -1 * df['returns']
df['neg_returns_lag6'] = df.groupby(
'code')['neg_returns'].shift(6)
df['ts_rank_neg'] = df.groupby(
'code')['neg_returns_lag6'].
transform(lambda x: ts_rank(x, 5)
)
df['rank_ts_rank'] = df.groupby('date')['ts_rank_neg'].
rank(pct=True)
comp3 = 0.73 * df['rank_ts_rank']

# Component 4: rank(abs(correlation(vwap, adv20, 6)))
df['corr_vwap_adv20'] = df.groupby('code').apply(
    lambda x: x['vwap'].rolling(6, min_periods=4).corr(x['adv20'])
).reset_index(level=0, drop=True)
df['abs_corr'] = df['corr_vwap_adv20'].abs()
df['rank_abs_corr'] = df.groupby('date')['abs_corr'].rank(
pct=True)
comp4 = df['rank_abs_corr']

# Component 5: 0.6 * rank((((sum(close, 200) / 200) - open) * (close - open)))
df['sum_close_200'] = df.groupby('code')['close'].
transform(
    lambda x: x.rolling(200, min_periods=150).sum()
)
df['sma200'] = df['sum_close_200'] / 200
df['sma200_open'] = df['sma200'] - df['open']
df['close_open2'] = df['close'] - df['open']
df['prod'] = df['sma200_open'] * df['close_open2']
df['rank_prod'] = df.groupby('date')['prod'].
rank(pct=True)
comp5 = 0.6 * df['rank_prod']

# 4. Combine all components
df['alpha'] = comp1 + comp2 + comp3 + comp4 + comp5

# 5. Pivot to desired format: index=date, columns=code, values=alpha
alpha_pivot = df.pivot_table(index='date', columns='code', values='alpha')

# 6. Output
print("Alpha factor pivot table shape:", alpha_pivot.shape)
print("\nFirst 5 rows and 5 columns:")
print(alpha_pivot.iloc[:5, :5])

alpha_pivot.to_csv("results.csv")
\end{lstlisting}


\end{document}